\documentclass[sigconf,nonacm]{acmart}
\usepackage{booktabs}
\usepackage{multirow}
\usepackage{amsmath}
\usepackage{bm}
\usepackage{graphicx}
\usepackage{xspace}
\usepackage{placeins}
\usepackage{float}
\usepackage{enumitem}

\setcopyright{none}
\renewcommand\footnotetextcopyrightpermission[1]{}

\newcommand{\method}{MDTE\xspace}

\begin{document}

\title{MDTE: Minority-Aware Diffusion over Temporal Edge Events for Imbalanced Node Classification}

\author{Zelong Zhou}
\affiliation{%
  \institution{Zhejiang University of Technology}
  \city{Hangzhou}
  \country{China}
}
\email{221125120233@zjut.edu.cn}

\author{Tianming Zhang}
\authornote{Corresponding author.}
\affiliation{%
  \institution{Zhejiang University of Technology}
  \city{Hangzhou}
  \country{China}
}
\email{tmzhang@zjut.edu.cn}

\author{Zhengyi Yang}
\affiliation{%
  \institution{The University of Sydney}
  \city{Sydney}
  \country{Australia}
}
\email{zhengyi.yang@sydney.edu.au}

\author{Yifu Tang}
\affiliation{%
  \institution{Vecton AI}
  \country{Australia}
}
\email{yves.tang@vectonai.com}

\author{Chenyu Hou}
\affiliation{%
  \institution{Zhejiang University of Technology}
  \city{Hangzhou}
  \country{China}
}
\email{houcy@zjut.edu.cn}

\author{Bin Cao}
\affiliation{%
  \institution{Zhejiang University of Technology}
  \city{Hangzhou}
  \country{China}
}
\email{bincao@zjut.edu.cn}

\author{Jing Fan}
\affiliation{%
  \institution{Zhejiang University of Technology}
  \city{Hangzhou}
  \country{China}
}
\email{fanjing@zjut.edu.cn}

\renewcommand{\shortauthors}{Zhou et al.}

\begin{abstract}
Class-imbalanced node classification on temporal graphs is challenging because majority-dominated temporal propagation progressively assimilates minority representations, while conventional node and neighborhood information provides insufficient discriminative evidence for minority classes. To address these issues, we propose \method, a minority-aware diffusion framework that reconstructs stable and discriminative temporal edge-event representations through conditional diffusion denoising. Specifically, \method introduces Distribution-Aware Selective Propagation, which combines Local Outlier Factor (LOF)-based propagation filtering with cluster-aware low-frequency propagation. The module preserves informative neighborhood dependencies while mitigating harmful propagation and majority-class information assimilation. It further develops Multi-View Discriminative Fusion, which exploits feature reconstruction and topology prediction to characterize class-wise differences in distribution learning and extracts complementary discriminability signals to guide denoising. Experiments on five real-world datasets demonstrate that \method consistently achieves the best performance on minority-class-oriented metrics, improving minority-class recall by up to 23.53 percentage points, minority-class F1 by 8.68 percentage points, and AUPRC by 2.67 percentage points over the strongest baselines.
\end{abstract}

\keywords{temporal graph learning, class-imbalanced node classification, graph diffusion models}

\maketitle

\section{Introduction}

Complex interaction networks, including financial transaction, blockchain transfer, and social networks, continuously evolve and can be naturally represented as temporal graphs of timestamped interactions~\cite{xu2020tgat,rossi2020tgn,yu2023dygformer}. Many real-world temporal graphs also exhibit class imbalance, where minority-class nodes, such as fraudulent accounts, illicit transaction nodes, and malicious users, are scarce but often of greater practical interest. Under temporal message propagation, their representations can be distorted by deviating neighboring representations and progressively assimilated toward majority-class patterns as majority information repeatedly dominates historical aggregation. Meanwhile, minority-specific patterns are underrepresented in the overall interaction distribution, providing insufficient discriminative evidence for separating minority and majority classes. Therefore, learning temporally stable and class-discriminative minority representations remains a challenge in class-imbalanced temporal graph learning.

In class-imbalanced temporal graphs, continuous interactions can progressively destabilize minority-class node representations, exacerbating the representational bias caused by class imbalance.
Most existing class-imbalanced graph learning methods are designed for static graphs and mainly alleviate classification bias through data-level and algorithm-level strategies~\cite{ma2025cilg}. In these methods, indiscriminate neighborhood aggregation may propagate representations that deviate from the local neighborhood distribution and introduce majority-class information into minority representations. We refer to such propagation as \emph{harmful propagation}. In temporal graphs, evolving interactions and continuous neighborhood propagation cause the effects of harmful propagation to accumulate and amplify over time, driving minority representations toward majority-class patterns. We refer to this process as \emph{majority-class information assimilation}.

Diffusion models can recover informative representations from perturbed inputs through denoising. Applying this denoising mechanism to graphs helps obtain more stable node representations. Existing graph diffusion methods have been explored for node representation learning~\cite{zhu2023ddm,chen2025graffe,zhu2025sdmg}, and recent studies have further extended diffusion to temporal graphs~\cite{tian2024conda,duc2026sdg}. These methods are generally optimized according to the overall graph distribution. Under class imbalance, majority-class patterns are more prevalent in the training distribution and can dominate the denoising objective, leaving insufficient signals for learning minority-specific patterns. Consequently, directly applying existing temporal diffusion models may recover globally plausible representations while failing to preserve minority-specific characteristics and sharpen inter-class boundaries. 

Effectively handling class-imbalanced node classification on temporal graphs requires simultaneously preserving the stability of minority-class representations and enhancing their class discriminability. This gives rise to two key technical challenges:

\textbf{Challenge I: How can temporal diffusion mitigate majority-class information assimilation while preserving informative neighborhood dependencies?}
In temporal graphs, repeated neighborhood interactions and message propagation cause the effects of harmful propagation to accumulate and amplify, gradually shifting minority representations toward majority-class patterns and thereby resulting in majority-class information assimilation. This process destabilizes minority representations and weakens minority-specific patterns. However, simply reducing the overall propagation strength may also discard informative neighborhood dependencies. Therefore, the key challenge is to \emph{mitigate majority-class information assimilation while preserving useful neighborhood information}, thereby maintaining the stability of minority representations.

\textbf{Challenge II: How can latent class-discriminative evidence be extracted to guide minority-aware denoising?}
The scarcity of minority interactions limits the discriminative information available for learning minority-specific distributions. Node representations alone may not sufficiently expose subtle class differences, while topology alone may overlook feature-level distinctions. A more effective diffusion process therefore requires complementary evidence from multiple views that can characterize how well the learned representations capture class-specific distributions. The key challenge is to \emph{extract and integrate complementary discriminability signals and incorporate them into the denoising process}, so that diffusion reconstruction is guided toward representations with clearer minority-majority separation.

To address these challenges, we propose \method, a \textbf{M}inority-aware \textbf{D}iffusion framework over \textbf{T}emporal \textbf{E}dge events for imbalanced node classification. \method represents timestamped interactions as temporal edge events and learns their distributions through forward diffusion and conditional reverse denoising. To address Challenge I, we propose \emph{Distribution-Aware Selective Propagation}, which provides propagation-aware denoising conditions through two complementary mechanisms: Local Outlier Factor (LOF)-based propagation filtering restricts the transmission of deviating representations, while cluster-aware low-frequency propagation suppresses cross-cluster information to reduce majority-class information assimilation without discarding useful neighborhood dependencies. To address Challenge II, we propose \emph{Multi-View Discriminative Fusion}, which performs feature reconstruction and topology prediction on original and low-frequency-propagated representations. By characterizing class-wise differences in distribution learning from feature and topology views, it extracts complementary discriminability signals and fuses them to guide the reconstruction of stable and class-discriminative temporal edge-event representations. Finally, the denoised edge-event representations are aggregated to the node level, followed by Debiased Contrastive Learning for downstream node classification.

In summary, our main contributions are as follows:

\begin{itemize}[leftmargin=*]
\item We propose \method, a minority-aware diffusion framework over temporal edge events that explicitly addresses the two fundamental issues of \emph{majority-class information assimilation} and \emph{insufficient class discriminability} in class-imbalanced temporal graphs. By performing condition-guided diffusion denoising on temporal edge events, \method learns representations that are both temporally stable and class-discriminative.
\item We develop two complementary mechanisms for constructing denoising conditions during reverse diffusion: \emph{Distribution-Aware Selective Propagation} and \emph{Multi-View Discriminative Fusion}. The former combines LOF-based propagation filtering and cluster-aware low-frequency propagation to suppress deviating representations and majority-class information assimilation while retaining useful neighborhood dependencies; the latter jointly exploits feature reconstruction and topology prediction to extract complementary discriminability signals and guide the reconstruction toward more class-discriminative temporal representations.
\item Extensive experiments on five real-world datasets demonstrate that \method consistently outperforms thirteen baseline methods, achieving particularly substantial improvements in minority-class recognition, with minority-class recall improved by up to \textbf{23.53} percentage points over the strongest baselines.
\end{itemize}

\section{Related Work}

\subsection{Class-Imbalanced Learning Methods}
Class imbalance is common in real-world classification tasks and is characterized by unequal sample sizes across classes. In graph node classification, minority-class nodes are scarce and sparsely labeled, while their representations can be dominated by information from majority-class neighbors, further exacerbating classification bias. Existing methods can be divided into data-level and algorithm-level approaches~\cite{ma2025cilg}.

Data-level methods include data interpolation~\cite{zhao2021graphsmote,park2022graphens,liu2022gatsmote,wu2021graphmixup}, adversarial generation~\cite{qu2021imgagn,duan2022sorag}, and pseudo-labeling~\cite{zhou2018sparc,zhou2023graphsr}. Among data-interpolation methods, GraphSMOTE~\cite{zhao2021graphsmote} generates minority-class nodes in the embedding space and predicts their structural connections, while GraphENS~\cite{park2022graphens} synthesizes complete ego-networks centered on minority-class nodes. Among adversarial-generation methods, ImGAGN~\cite{qu2021imgagn} generates minority-class nodes and their connections through adversarial learning, while SORAG~\cite{duan2022sorag} extends this approach to imbalanced multi-label graphs. Among pseudo-labeling methods, SPARC~\cite{zhou2018sparc} combines label propagation with self-paced learning to select minority-class pseudo-labels, while GraphSR~\cite{zhou2023graphsr} uses node similarity and reinforcement learning to improve pseudo-label reliability.

Algorithm-level methods include model refinement~\cite{wang2018rsdne,wang2022egcn,shi2020drgcn}, loss-function design~\cite{chen2021renode,song2022tam}, and long-tailed representation enhancement~\cite{yun2022lte4g}. Among model-refinement methods, RSDNE~\cite{wang2018rsdne} constrains node embeddings using intra-class similarity and inter-class dissimilarity, EGCN~\cite{wang2022egcn} adjusts cross-class neighborhood aggregation from local and global perspectives, and DR-GCN~\cite{shi2020drgcn} aligns the latent representation distributions of labeled and unlabeled nodes. Among loss-function methods, ReNode~\cite{chen2021renode} adjusts training weights according to node positions relative to topological class boundaries, while TAM~\cite{song2022tam} adjusts classification margins according to local neighborhood label distributions. For long-tailed representation enhancement, LTE4G~\cite{yun2022lte4g} combines expert models, knowledge distillation, and class prototypes to address class-level and node-degree-level long tails.

However, most existing methods are designed for static graphs and therefore do not explicitly capture interaction order or the accumulation of historical information. They also do not adequately address propagation interference from deviating representations or the repeated influence of majority-class information during temporal aggregation, thereby progressively destabilizing minority-class representations. Moreover, the discriminative information extracted by these methods remains insufficient to further distinguish minority-class nodes from majority-class nodes.

\subsection{Graph Denoising Diffusion Models}
Graph denoising diffusion models learn graph representations or graph data distributions over node features, graph structures, or latent representations through forward perturbation and reverse denoising. Existing static-graph methods mainly include representation-space diffusion~\cite{zhu2023ddm,chen2025graffe,zhu2025sdmg}, latent-space diffusion~\cite{zhou2024lgd,dai2024diffgcl}, and structure-oriented diffusion~\cite{vignac2023digress,wesego2025ddgae}.

In representation space, DDM~\cite{zhu2023ddm} introduces data-dependent directional noise to learn semantically informative graph representations, Graffe~\cite{chen2025graffe} uses encoder-produced graph representations to guide node-feature recovery, and SDMG~\cite{zhu2025sdmg} combines low-frequency encoding with multi-scale smoothing constraints. In latent space, LGD~\cite{zhou2024lgd} performs diffusion over latent graph representations to unify graph generation and prediction. DiffGCL~\cite{dai2024diffgcl} diffuses low-dimensional latent embeddings produced by a shared graph encoder and combines diffusion with graph contrastive learning. For graph structures, DiGress~\cite{vignac2023digress} diffuses discrete node and edge types, while DDGAE~\cite{wesego2025ddgae} reconstructs graph structures through a discrete diffusion decoder.

Recently, a small number of studies have extended diffusion mechanisms to continuous-time dynamic graphs~\cite{tian2024conda,duc2026sdg}: Conda~\cite{tian2024conda} generates historical neighbor embeddings through latent conditional diffusion for dynamic graph augmentation, while SDG~\cite{duc2026sdg} reconstructs historical interaction sequences through conditional denoising for temporal link prediction.

However, existing diffusion methods for continuous-time dynamic graphs do not explicitly address class imbalance and lack targeted modeling of minority-class node representations. Majority-class interactions tend to dominate diffusion training, biasing the model toward recovering the majority-class data distribution. Meanwhile, the limited occurrence of minority-class patterns makes them difficult to learn adequately. Denoising and propagation can further weaken the distinctive characteristics of minority-class representations, thereby blurring inter-class boundaries and degrading minority-class classification performance.

\section{Preliminaries}

\noindent\textbf{Definition 3.1 (Continuous-time Temporal Subgraph).}
Given a batch of temporal edge events observed up to time $T$, the induced temporal subgraph is denoted by $\mathcal{G}^{T}=(\mathcal{V}^{T},\mathcal{E}^{T},\mathbf{A}^{T},\mathbf{H})$, where $\mathcal{V}^{T}=\{v_1,v_2,\ldots,v_N\}$ is the set of nodes involved in the batch up to time $T$, and $N=|\mathcal{V}^{T}|$. The temporal edge-event set is $\mathcal{E}^{T}=\{e_{ij}^{t}=(v_i,v_j,t)\mid v_i,v_j\in\mathcal{V}^{T},\,t\leq T\}$, where $e_{ij}^{t}$ denotes an interaction between $v_i$ and $v_j$ at time $t$; the same node pair can interact multiple times. $\mathbf{A}^{T}$ is the adjacency matrix constructed from the interactions in $\mathcal{E}^{T}$, and $\mathbf{H}=[\mathbf{h}_1,\ldots,\mathbf{h}_N]^{\top}\in\mathbb{R}^{N\times d}$ is the node-representation matrix, where $\mathbf{h}_i$ is the $d$-dimensional representation of $v_i$.

\par\smallskip
\noindent\textbf{Definition 3.2 (Continuous-time Encoding).}
For an edge event occurring at time $t$, let $\Delta t=T-t$ denote the interval between its occurrence time and the observation cutoff time. The continuous-time encoding is generated using multi-scale sine and cosine basis functions~\cite{xu2020tgat}:
\[
\begin{aligned}
\boldsymbol{\Phi}_{\mathrm{time}}(\Delta t)
={}&\sqrt{\frac{1}{2d_T}}\bigl[
\cos(\omega_1\Delta t),\sin(\omega_1\Delta t),\ldots,\\[-2pt]
&\cos(\omega_{d_T}\Delta t),\sin(\omega_{d_T}\Delta t)
\bigr]^{\top}.
\end{aligned}
\]
Here, $d_T$ is the number of frequencies, and $\boldsymbol{\omega}=(\omega_1,\ldots,\omega_{d_T})^{\top}$ contains learnable frequency parameters. This encoding represents the temporal position of an edge event relative to the observation cutoff time at multiple temporal scales.

\par\smallskip
\noindent\textbf{Definition 3.3 (Random-Walk Positional Encoding).}
Given a graph with adjacency matrix $\mathbf{A}$, the random-walk transition matrix is defined as $\mathbf{P}=\mathbf{D}^{-1}\mathbf{A}$, where $\mathbf{D}$ is the degree matrix. For node $v_i$, the random-walk positional encoding is defined as $\mathbf{p}_i=[(\mathbf{P})_{ii}\Vert(\mathbf{P}^{2})_{ii}\Vert\cdots\Vert(\mathbf{P}^{K_{\mathrm{rw}}})_{ii}]$, where $(\mathbf{P}^{s})_{ii}$ denotes the probability that a random walk starting from $v_i$ returns to itself after $s$ steps, and $K_{\mathrm{rw}}$ is the maximum walk length. These return probabilities characterize the structural position of a node across different neighborhood ranges~\cite{rampasek2022recipe}.

\par\smallskip
\noindent\textbf{Definition 3.4 (Imbalanced Temporal Graph Node Classification).}
Given the continuous-time temporal subgraph $\mathcal{G}^{T}$ and a set of labeled nodes $\mathcal{V}_{L}$ with an imbalanced class distribution, partitioned into training, validation, and test sets, the objective is to learn a node-classification function under class imbalance, $f_{\Theta}:(v_i,\mathcal{E}^{T},\mathbf{A}^{T},\mathbf{H})\mapsto\widehat{y}_{i}^{T}$. Here, $\Theta$ denotes the learnable parameters, and $\widehat{y}_{i}^{T}$ is the predicted label of node $v_i$ at time $T$. The function predicts node classes from node representations and historical interactions up to time $T$, with particular emphasis on accurately identifying minority-class nodes.

\section{Methodology}
\label{sec:method}

\begin{figure*}
  \centering
  \includegraphics[width=\textwidth]{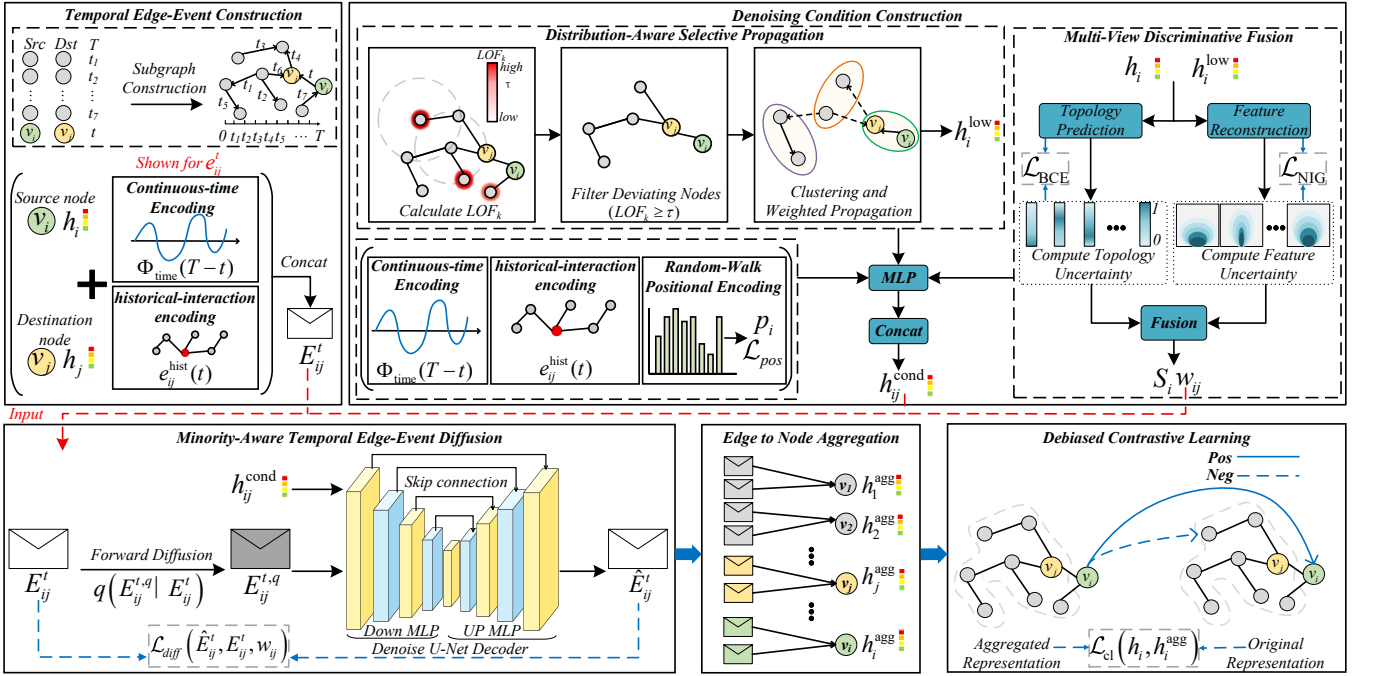}
  \caption{Overall architecture of \method.}
  \label{fig:framework}
\end{figure*}

To address propagation interference from deviating representations, majority-class information assimilation, and insufficient discriminative information, we propose \method, a minority-aware diffusion framework over temporal edge events for imbalanced node classification. As illustrated in Figure~\ref{fig:framework}, \method consists of four stages: Temporal Edge-Event Construction, Minority-Aware Temporal Edge-Event Diffusion, Edge-to-Node Aggregation, and Debiased Contrastive Learning. The second stage proceeds through Forward Diffusion, Denoising Condition Construction, and Conditional Reverse Process. The denoising condition is constructed by two key modules: Distribution-Aware Selective Propagation and Multi-View Discriminative Fusion.

\subsection{Temporal Edge-Event Construction}

Existing methods for class-imbalanced graph learning and graph diffusion typically aggregate historical interactions into a fixed graph structure, making it difficult to preserve the temporal information of individual interactions and the endpoint states before they occur. Therefore, \method represents each timestamped node interaction as a temporal edge event.

For a temporal edge event $e_{ij}^{t}=(v_i,v_j,t)$, its representation is
\begin{equation}
\mathbf{E}_{ij}^{t}=\left[\mathbf{h}_{i}\Vert\mathbf{h}_{j}\Vert\boldsymbol{\Phi}_{\mathrm{time}}(\Delta t)\Vert\mathbf{e}_{ij}^{\mathrm{hist}}(t)\right],
\end{equation}
where $\mathbf{h}_{i}$ and $\mathbf{h}_{j}$ are the representations of the interacting endpoints, $\boldsymbol{\Phi}_{\mathrm{time}}(\Delta t)$ is the continuous-time encoding, $\mathbf{e}_{ij}^{\mathrm{hist}}(t)$ encodes the endpoints' historical interactions before the event, and $\Vert$ denotes vector concatenation.

Let $d_i(t^-)$ and $d_j(t^-)$ be the numbers of historical interactions involving nodes $v_i$ and $v_j$, respectively, before time $t$. The historical-interaction encoding is
\begin{equation}
\mathbf{e}_{ij}^{\mathrm{hist}}(t)=\mathrm{MLP}
\left(\log\left(1+d_i(t^-)+d_j(t^-)\right)\right).
\end{equation}
The logarithmic transformation reduces the excessive influence of large interaction counts, and the resulting $\mathbf{E}_{ij}^{t}$ serves as the diffusion object in the subsequent edge-event diffusion process.

\subsection{Minority-Aware Temporal Edge-Event Diffusion}

To learn stable edge-event representations while reducing majority-class bias, \method adopts conditional diffusion with controlled noise and minority-oriented denoising conditions.

\subsubsection{Forward Diffusion}

To enable the conditional reverse process to learn the distribution of temporal edge-event representations, we add controlled noise to $\mathbf{E}_{ij}^{t}$. During the forward process~\cite{ho2020denoising}, we randomly sample a diffusion step $q\in\{1,\ldots,Q\}$, where $Q$ denotes the total number of diffusion steps, and add directional noise $\epsilon_{ij}$ constructed from the distributional characteristics of the edge-event representation~\cite{zhu2023ddm} to obtain
\begin{equation}
\mathbf{E}_{ij}^{t,q}=\sqrt{\bar{\alpha}_q}\,\mathbf{E}_{ij}^{t}+\sqrt{1-\bar{\alpha}_q}\,\epsilon_{ij},
\end{equation}
\begin{equation}
\bar{\alpha}_q=\prod_{r=1}^{q}\alpha_r.
\end{equation}
Here, $\alpha_r\in(0,1)$ denotes the information-retention coefficient at diffusion step $r$, and $\bar{\alpha}_q$ represents the cumulative proportion of original edge-event information retained after $q$ diffusion steps. Thus, $\sqrt{\bar{\alpha}_q}$ controls the proportion of the original representation, whereas $\sqrt{1-\bar{\alpha}_q}$ controls the proportion of noise.

\subsubsection{Denoising Condition Construction}

To mitigate propagation interference and majority-class information assimilation while extracting additional discriminative information, we propose Distribution-Aware Selective Propagation and Multi-View Discriminative Fusion to construct the denoising conditions.

\paragraph{Distribution-Aware Selective Propagation}

Some node representations deviate from their neighborhood distribution and interfere with other nodes during propagation. Therefore, we propose Distribution-Aware Selective Propagation, which uses LOF to limit their propagation to other nodes while preserving their own information through self-loops~\cite{breunig2000lof}.

We compute LOF over the nodes in $\mathcal{G}^{T}$ to quantify how strongly each node representation deviates from its neighborhood distribution. Given the $k$-nearest-neighbor set $\mathcal{N}_k(i)$ of node $v_i$, its reachability distance to neighbor $v_j$ is
\begin{equation}
\label{eq:reachability-distance}
\mathrm{rd}_k(i,j)=\max\left(k\text{-}\mathrm{dist}(j),\mathrm{dist}(i,j)\right),
\end{equation}
where $k\text{-}\mathrm{dist}(j)$ is the distance from $v_j$ to its $k$-th nearest neighbor, and $\mathrm{dist}(i,j)$ is the cosine distance between $v_i$ and $v_j$ in the representation space. The reachability distance uses the local neighborhood radius $k\text{-}\mathrm{dist}(j)$ of $v_j$ as a lower bound, preventing overly small pairwise distances in locally dense regions from exerting excessive influence on density estimation and thereby improving its stability.

Based on the reachability distance, the local reachability density of $v_i$ is
\begin{equation}
\mathrm{lrd}_k(i)=\left(
\frac{\sum_{v_j\in\mathcal{N}_k(i)}\mathrm{rd}_k(i,j)}{|\mathcal{N}_k(i)|}
\right)^{-1}.
\end{equation}
The local reachability density is the inverse of the average reachability distance between a node and its neighbors and measures local compactness; a larger $\mathrm{lrd}_k(i)$ indicates a denser local region.

The local outlier factor of $v_i$ is
\begin{equation}
\mathrm{LOF}_k(i)=\frac{1}{|\mathcal{N}_k(i)|}
\sum_{v_j\in\mathcal{N}_k(i)}\frac{\mathrm{lrd}_k(j)}{\mathrm{lrd}_k(i)}.
\end{equation}
LOF measures how strongly a node representation deviates from its neighborhood by comparing the node's local reachability density with those of its neighbors. When the local reachability density of $v_i$ is substantially lower than those of its neighbors, $\mathrm{LOF}_k(i)$ increases, indicating a stronger deviation from the dominant representation pattern of its neighborhood.

We then rank-normalize the logarithmic values of $\mathrm{LOF}_k(i)$ to obtain relative deviation scores $u_i\in[0,1]$. Given a propagation-filtering threshold $\tau$, we construct a node mask $\mathbf{m}\in\{0,1\}^{N}$, whose $i$-th entry is set to 1 if $u_i<\tau$ and 0 otherwise. The filtered adjacency matrix is
\begin{equation}
\label{eq:filtered-adjacency}
\widetilde{\mathbf{A}}^{T}=\mathbf{A}^{T}\odot(\mathbf{m}\mathbf{m}^{\top})+\mathbf{I}.
\end{equation}
Here, $\mathbf{I}$ is the identity matrix that retains a self-loop for every node, and $\odot$ denotes element-wise multiplication.

Cross-cluster propagation mixes dissimilar representation patterns and amplifies majority-dominated information. Therefore, we propose cluster-aware low-frequency propagation to preserve intra-cluster propagation while downweighting cross-cluster messages.

We apply K-means~\cite{lloyd1982least} to the nodes retained after propagation filtering and use the elbow method~\cite{thorndike1953family} to adaptively determine the number of clusters $K_{\mathrm{c}}$. The elbow method selects the point at which the decrease in within-cluster inertia changes from rapid to gradual, where the inertia is the mean squared Euclidean distance from each node to its assigned cluster center and measures cluster compactness. Let $c_i\in\{1,\ldots,K_{\mathrm{c}}\}$ denote the resulting cluster label of node $v_i$. The propagation weight is
\begin{equation}
B_{ij}=\begin{cases}
1,&c_i=c_j,\\
\alpha,&c_i\neq c_j,
\end{cases}
\qquad 0<\alpha<1,
\end{equation}
where $\alpha$ is a learnable cross-cluster propagation weight. The weighted adjacency matrix for low-frequency propagation is
\begin{equation}
\overline{A}_{ij}^{T}=\widetilde{A}_{ij}^{T}B_{ij}.
\end{equation}
Following the Node Feature Low-Frequency Information Encoder in SDMG~\cite{zhu2025sdmg}, we extend it to the weighted filtered graph, where the cluster-aware edge weights modulate the GAT attention coefficients. The resulting low-frequency node representations are
\begin{equation}
\mathbf{H}^{\mathrm{low}}=\mathrm{LP}(\overline{\mathbf{A}}^{T},\mathbf{H}).
\end{equation}
Here, $\mathrm{LP}(\cdot)$ denotes the GAT-based low-frequency encoder, $\mathbf{H}^{\mathrm{low}}=[\mathbf{h}_1^{\mathrm{low}},\ldots,\mathbf{h}_N^{\mathrm{low}}]^{\top}$, and $\mathbf{h}_i^{\mathrm{low}}$ is the low-frequency representation of node $v_i$.

For a minority-class node, same-cluster neighbors usually exhibit similar representation patterns, and preserving within-cluster propagation captures smooth neighborhood information. In contrast, cross-cluster neighbors may introduce inconsistent information due to larger representation discrepancies. Therefore, downweighting cross-cluster propagation preserves useful information from similar neighbors while limiting dissimilar majority-class information, thereby alleviating majority-class information assimilation.

\paragraph{Multi-View Discriminative Fusion}

To further distinguish minority-class nodes from majority-class nodes, we propose Multi-View Discriminative Fusion, which extracts and fuses feature-based and topology-based discriminability signals from the original and low-frequency node representations~\cite{wei2025gel}.

The attribute patterns and connection relations of majority-class nodes are observed more frequently during training, making their distribution easier for the model to learn. In contrast, minority-class nodes and their associated interactions are limited, so their distinctive patterns can remain insufficiently learned. Based on this difference in how well the underlying distributions are learned, we extract discriminability signals through feature reconstruction and topology prediction to capture additional differences between minority-class nodes and majority-class nodes.

\noindent\textbf{Feature Discriminability Signal.}
Directly reconstructing the complete node representation yields only an overall reconstruction error, whereas dimension-wise Normal-Inverse-Gamma (NIG) modeling estimates the distributional uncertainty of individual dimensions, providing finer-grained discriminative information for characterizing feature differences between minority-class and majority-class nodes. Specifically, we model each representation dimension as a Gaussian variable with unknown mean and variance and use an NIG distribution~\cite{amini2020deep} to describe parameter uncertainty. Given the original node representation $\mathbf{h}_i$, the feature signal head outputs four groups of NIG parameters:
\begin{equation}
(\boldsymbol{\gamma}_i,\boldsymbol{\kappa}_i,
\boldsymbol{\alpha}_i,\boldsymbol{\beta}_i)
=\operatorname{MLP}(\mathbf{h}_i).
\end{equation}
The four output vectors contain a set of NIG parameters for each dimension of the node representation. For dimension $\ell$, $\gamma_{i,\ell}$ is the reconstructed value of $h_{i,\ell}$; $\kappa_{i,\ell}$ reflects the certainty of the reconstructed value, with a larger value indicating higher certainty; and $\alpha_{i,\ell}$ and $\beta_{i,\ell}$ control the shape and scale of the variance distribution, respectively. Together, $\kappa_{i,\ell}$, $\alpha_{i,\ell}$, and $\beta_{i,\ell}$ characterize the reconstruction uncertainty of dimension $\ell$. For each dimension, $\kappa_{i,\ell}>0$, $\alpha_{i,\ell}>1$, and $\beta_{i,\ell}>0$.

The feature discriminability signal of node $v_i$ is derived from the reconstruction uncertainty:
\begin{equation}
s_i=\operatorname*{Mean}\left(
\sqrt{\frac{\beta_{i,\ell}}
{\kappa_{i,\ell}(\alpha_{i,\ell}-1)}}
\right).
\end{equation}
The quantity inside the square root is the reconstruction uncertainty in dimension $\ell$, and $\operatorname*{Mean}(\cdot)$ averages over all representation dimensions. A larger $s_i$ indicates higher uncertainty about reconstructing the node's feature pattern. Under class imbalance, minority-class feature patterns occur less frequently and are more difficult to reconstruct stably; therefore, a node with a larger $s_i$ is more likely to exhibit a minority-class feature pattern.

To obtain reliable reconstruction uncertainty, we use $\mathcal{L}_{\mathrm{NLL}}$ to fit the predictive distribution to the original node representation and $\mathcal{L}_{\mathrm{reg}}$ to prevent overconfidence when the reconstruction error is large. The NIG objective is
\begin{equation}
\mathcal{L}_{\mathrm{NIG}}
=\mathcal{L}_{\mathrm{NLL}}
+\lambda_e\mathcal{L}_{\mathrm{reg}},
\end{equation}
where $\lambda_e$ controls the contribution of the regularization term.

Following evidential regression~\cite{amini2020deep}, the negative log-likelihood for node $v_i$ is
\begin{align}
\mathcal{L}_{\mathrm{NLL},i}=\operatorname*{Mean}\Bigg[&
\frac{1}{2}\log\frac{\pi}{\kappa_{i,\ell}}
-\alpha_{i,\ell}\log\left(2\beta_{i,\ell}(1+\kappa_{i,\ell})\right) \nonumber\\
&+\left(\alpha_{i,\ell}+\frac{1}{2}\right)
\log\left(\kappa_{i,\ell}(h_{i,\ell}-\gamma_{i,\ell})^2
+2\beta_{i,\ell}(1+\kappa_{i,\ell})\right) \nonumber\\
&+\log\Gamma(\alpha_{i,\ell})
-\log\Gamma\left(\alpha_{i,\ell}+\frac{1}{2}\right)\Bigg].
\end{align}
The squared reconstruction deviation $(h_{i,\ell}-\gamma_{i,\ell})^2$ appears in the logarithmic residual term. A larger deviation increases the loss and encourages $\gamma_{i,\ell}$ to approach $h_{i,\ell}$. Meanwhile, $\kappa_{i,\ell}$, $\alpha_{i,\ell}$, and $\beta_{i,\ell}$ adjust the penalty according to the estimated uncertainty: for the same reconstruction deviation, lower uncertainty results in a stronger penalty. Consequently, minimizing $\mathcal{L}_{\mathrm{NLL},i}$ jointly promotes accurate reconstruction and learns uncertainty consistent with the reconstruction deviation.

For node $v_i$, the NIG regularization term is
\begin{equation}
\mathcal{L}_{\mathrm{reg},i}
=\operatorname*{Mean}\left[
|h_{i,\ell}-\gamma_{i,\ell}|
(2\kappa_{i,\ell}+\alpha_{i,\ell})
\right],
\end{equation}
A larger reconstruction deviation $\lvert h_{i,\ell}-\gamma_{i,\ell}\rvert$, together with a larger value of $2\kappa_{i,\ell}+\alpha_{i,\ell}$, results in a stronger penalty, thereby encouraging higher uncertainty for inaccurately reconstructed feature patterns. Both $\mathcal{L}_{\mathrm{NLL}}$ and $\mathcal{L}_{\mathrm{reg}}$ are obtained by averaging their node-level terms over the current batch.

\noindent\textbf{Topology Discriminability Signal.}
Feature-reconstruction uncertainty reflects how well node attributes are learned but does not capture discriminative information in connection patterns. Therefore, from the topology view, a topology signal head predicts whether a connection exists between a pair of nodes. For a node pair $(v_i,v_j)$, the connection probability is
\begin{equation}
\epsilon_{ij}^{+}
=\operatorname{MLP}([\mathbf{h}_i\Vert\mathbf{h}_j]).
\end{equation}
The output layer of the MLP uses a sigmoid activation to map its output to $(0,1)$, and the probability that no connection exists is $\epsilon_{ij}^{-}=1-\epsilon_{ij}^{+}$.

Topology prediction uses observed connections as positive samples and randomly sampled unconnected node pairs as negative samples. For a node pair $(v_i,v_j)$, the binary cross-entropy loss is
\begin{equation}
\mathcal{L}_{ij}^{\mathrm{BCE}}
=-\left[A_{ij}^{T}\log\epsilon_{ij}^{+}
+(1-A_{ij}^{T})\log\epsilon_{ij}^{-}\right],
\end{equation}
where $A_{ij}^{T}=1$ for an observed connection and $A_{ij}^{T}=0$ for a sampled unconnected pair. The uncertainty of the connection prediction is measured using predictive entropy:
\begin{equation}
\eta_{ij}^{\mathrm{topo}}
=-\epsilon_{ij}^{+}\log\epsilon_{ij}^{+}
-\epsilon_{ij}^{-}\log\epsilon_{ij}^{-}.
\end{equation}
The topology discriminability signal combines connection-prediction error and prediction uncertainty:
\begin{equation}
u_{ij}^{\mathrm{topo}}
=\frac{1}{2}\left(
\mathcal{L}_{ij}^{\mathrm{BCE}}
+\eta_{ij}^{\mathrm{topo}}
\right).
\end{equation}
A larger $\mathcal{L}_{ij}^{\mathrm{BCE}}$ indicates greater prediction error, whereas a larger $\eta_{ij}^{\mathrm{topo}}$ indicates that the connection is predicted with high uncertainty. Under class imbalance, node pairs with larger prediction errors or uncertainties are more likely to correspond to insufficiently learned minority-class connection patterns.

Let $\Omega_{\mathrm{topo}}$ denote the positive and negative node pairs sampled for topology prediction in the current batch. For each node $v_i$, the topology discriminability signals associated with $v_i$ are averaged to obtain the node-level topology discriminability signal
\begin{equation}
s_i^{\mathrm{topo}}
=\operatorname*{Mean}_{(v_i,v_j)\in\Omega_{\mathrm{topo}}}
\left(u_{ij}^{\mathrm{topo}}\right).
\end{equation}
To extract more discriminative signals and further distinguish minority-class nodes from majority-class nodes, we further propose extracting feature-based and topology-based discriminability signals from the low-frequency representations and fusing them with those obtained from the original representations.

The original node representation $\mathbf{h}_i$ mainly retains node-attribute information, whereas $\mathbf{h}_i^{\mathrm{low}}$ incorporates smoothed structural information from the filtered neighborhood. We apply the same feature-reconstruction and topology-prediction processes to both representation spaces using signal heads with the same structure but independent parameters.

To measure whether the propagated representation preserves node-specific patterns, we reconstruct $\mathbf{h}_i$ from $\mathbf{h}_i^{\mathrm{low}}$ and use the reconstruction uncertainty as the discriminability signal $s_i^{\mathrm{low}}$, with the corresponding loss $\mathcal{L}_{\mathrm{NIG}}^{\mathrm{low}}$. The joint feature objective is
\begin{equation}
\mathcal{L}_{\mathrm{feat}}
=\mathcal{L}_{\mathrm{NIG}}
+\mathcal{L}_{\mathrm{NIG}}^{\mathrm{low}}.
\end{equation}

In the low-frequency topology view, we use the low-frequency node representations to predict the same positive and negative node pairs. This yields the low-frequency topology discriminability signal $u_{ij}^{\mathrm{topo,low}}$ and the corresponding node-level signal $s_i^{\mathrm{topo,low}}$. The joint topology objective for the original and low-frequency views is
\begin{equation}
\mathcal{L}_{\mathrm{topo}}=
\operatorname*{Mean}_{(v_i,v_j)\in\Omega_{\mathrm{topo}}}
\left(\mathcal{L}_{ij}^{\mathrm{BCE}}+\mathcal{L}_{ij}^{\mathrm{BCE,low}}\right).
\end{equation}
To jointly learn the feature-based and topology-based discriminability signals from the original and low-frequency representation spaces, we combine the feature and topology objectives as
\begin{equation}
\mathcal{L}_{\mathrm{aux}}=\lambda_{\mathrm{feat}}\mathcal{L}_{\mathrm{feat}}+\lambda_{\mathrm{topo}}\mathcal{L}_{\mathrm{topo}},
\end{equation}
where $\lambda_{\mathrm{feat}}$ and $\lambda_{\mathrm{topo}}$ control the contributions of the feature and topology objectives, respectively.

To provide complementary discriminative information for further distinguishing minority-class nodes from majority-class nodes, we concatenate the feature-based and topology-based discriminability signals from the original and low-frequency representation spaces:
\begin{equation}
\mathbf{S}_i=
[s_i\Vert s_i^{\mathrm{low}}\Vert
s_i^{\mathrm{topo}}\Vert s_i^{\mathrm{topo,low}}].
\end{equation}
The resulting signal $\mathbf{S}_i$ integrates feature-reconstruction uncertainty and topology-prediction error and uncertainty from the two representation spaces.

The mean of the four signals in $\mathbf{S}_i$ is used as the node-level discriminability weight $w_i$. For each temporal edge event $e_{ij}^{t}$, the corresponding edge-event weight $w_{ij}$ is obtained by averaging the weights of its two endpoints.

\paragraph{Condition Integration}

To construct an informative denoising condition that more effectively guides the reverse process, \method integrates the low-frequency node representation $\mathbf{h}_i^{\mathrm{low}}$ and fused discriminability signal $\mathbf{S}_i$, while incorporating the random-walk positional representation $\mathbf{p}_i$ to complement structural-position information. To maintain consistency between the random-walk positional representations and the filtered propagation topology, we apply standard graph-Laplacian smoothness regularization~\cite{belkin2006manifold}, denoted by $\mathcal{L}_{\mathrm{pos}}$.

The resulting node-level denoising condition is
\begin{equation}
\widetilde{\mathbf{h}}_i=\mathrm{MLP}\left([\mathbf{h}_i^{\mathrm{low}}+\mathbf{p}_i\Vert\mathbf{S}_i]\right).
\end{equation}
To further incorporate event-specific temporal and historical information, we then combine the node-level conditions of the two endpoints with the temporal and historical-interaction encodings to construct the edge-event-level denoising condition:
\begin{equation}
\begin{aligned}
\mathbf{h}_{ij}^{\mathrm{cond}}={}&
[\mathbf{W}_{s}\widetilde{\mathbf{h}}_i
\Vert\mathbf{W}_{d}\widetilde{\mathbf{h}}_j
\Vert\mathbf{W}_{t}\boldsymbol{\Phi}_{\mathrm{time}}(\Delta t)
\Vert\mathbf{W}_{g}\mathbf{e}_{ij}^{\mathrm{hist}}(t)].
\end{aligned}
\end{equation}
Here, $\mathbf{W}_{s}$, $\mathbf{W}_{d}$, $\mathbf{W}_{t}$, and $\mathbf{W}_{g}$ are the respective projection matrices.

\subsubsection{Conditional Reverse Process}

To learn the distribution of stable and discriminative edge-event representations, \method performs conditional reverse denoising guided by $\mathbf{h}_{ij}^{\mathrm{cond}}$. Given $\mathbf{E}_{ij}^{t,q}$, diffusion step $q$, and $\mathbf{h}_{ij}^{\mathrm{cond}}$, a conditional U-Net reconstructs the edge-event representation:
\begin{equation}
\widehat{\mathbf{E}}_{ij}^{t}=\mathrm{U\mbox{-}Net}
(\mathbf{E}_{ij}^{t,q},q,\mathbf{h}_{ij}^{\mathrm{cond}}).
\end{equation}
To prevent frequent majority-class interactions from dominating diffusion training, the discriminability weight $w_{ij}$ is used to weight the squared reconstruction error of each edge event. The diffusion loss is
\begin{equation}
\mathcal{L}_{\mathrm{diff}}=
\frac{\sum_{e_{ij}^{t}\in\mathcal{B}_e}w_{ij}\left\|\widehat{\mathbf{E}}_{ij}^{t}-\mathbf{E}_{ij}^{t}\right\|_2^2}
{\sum_{e_{ij}^{t}\in\mathcal{B}_e}w_{ij}}.
\end{equation}
Here, $\mathcal{B}_e$ denotes the edge events in the current batch. A larger $w_{ij}$ increases the contribution of the corresponding edge event to diffusion training, thereby reducing majority-dominated learning.

\subsection{Edge-to-Node Aggregation and Debiased Contrastive Learning}

To obtain node representations for downstream classification while preserving edge-level information, \method combines each denoised edge-event representation with its temporal encoding, historical-interaction encoding, and source/destination role encoding after dimensional alignment, and aggregates the resulting representations by attention:
\begin{equation}
\mathbf{h}_i^{\mathrm{agg}}
=\operatorname{Attn}_{e_{ij}^{t}\in\mathcal{R}(i)}\left(
\widehat{\mathbf{E}}_{ij}^{t}
+\boldsymbol{\Phi}_{\mathrm{time}}(\Delta t)
+\mathbf{e}_{ij}^{\mathrm{hist}}(t)
+\mathbf{r}_{i\mid ij}^{t}
\right).
\end{equation}
Here, $\mathcal{R}(i)$ denotes the temporal edge events involving $v_i$ in the current batch, and $\mathbf{r}_{i\mid ij}^{t}$ is the role encoding that indicates whether $v_i$ is the source or destination of $e_{ij}^{t}$.

To maintain consistency between the denoised aggregated representation $\mathbf{h}_i^{\mathrm{agg}}$ and the original node representation $\mathbf{h}_i$, \method treats them as a positive pair and uses the original representations of other nodes in the current batch as negatives. Since these negatives may contain semantically similar false negatives, we adopt a simplified correction inspired by Debiased Contrastive Learning~\cite{chuang2020debiased}. The corrected negative term is
\begin{equation}
\label{eq:debiased-negative}
n_i=
\frac{
\displaystyle\sum_{j=1,\,j\neq i}^{N}
\exp\left(\mathrm{sim}(\mathbf{h}_i^{\mathrm{agg}},\mathbf{h}_j)/\tau_c\right)
-\tau_{+}\exp\left(\mathrm{sim}(\mathbf{h}_i^{\mathrm{agg}},\mathbf{h}_i)/\tau_c\right)
}{1-\tau_{+}}.
\end{equation}
Here, $\mathrm{sim}(\cdot,\cdot)$ denotes cosine similarity, $\tau_c$ is the temperature coefficient, and $\tau_{+}$ is the prior proportion of potential false negatives. The debiased contrastive loss is
\begin{equation}
\mathcal{L}_{\mathrm{cl}}
=-\frac{1}{N}\sum_{i=1}^{N}
\log
\frac{
\exp\left(\mathrm{sim}(\mathbf{h}_i^{\mathrm{agg}},\mathbf{h}_i)/\tau_c\right)
}{
\exp\left(\mathrm{sim}(\mathbf{h}_i^{\mathrm{agg}},\mathbf{h}_i)/\tau_c\right)+n_i
}.
\end{equation}

If a node appears in multiple edge-event batches, its denoised aggregated representations are averaged to obtain a unified node representation $\mathbf{h}_i^{\mathrm{node}}$ for downstream classification.

Combining the diffusion, auxiliary, positional, and debiased contrastive objectives, the overall training objective of \method is
\begin{equation}
\mathcal{L}_{\mathrm{total}}=
\mathcal{L}_{\mathrm{diff}}+
\lambda_{\mathrm{cl}}\mathcal{L}_{\mathrm{cl}}+
\mathcal{L}_{\mathrm{aux}}+
\lambda_{\mathrm{pos}}\mathcal{L}_{\mathrm{pos}}.
\end{equation}
The coefficients $\lambda_{\mathrm{cl}}$ and $\lambda_{\mathrm{pos}}$ control the debiased contrastive loss and positional-encoding constraint, respectively. This process does not use node class labels; after training, the parameters of \method are frozen, and a unified downstream classifier is trained using labeled training nodes.

\section{Experiments}
\nopagebreak[4]

\subsection{Experimental Setup}

\noindent\textbf{Datasets.}
We conduct experiments on five financial transaction graph datasets: DGraph-Fin~\cite{huang2022dgraph}, Elliptic~\cite{weber2019elliptic}, Elliptic++ Transactions and Actors~\cite{elmougy2023ellipticpp}, and Ethereum~\cite{chen2021phishing}. Table~\ref{tab:datasets} reports their statistics, where the imbalance ratio (IR) is the ratio of majority- to minority-class nodes; a larger IR indicates more severe class imbalance. DGraph-Fin identifies fraudulent users, Elliptic and Elliptic++ Transactions detect illicit transactions, Elliptic++ Actors identifies illicit participants, and Ethereum detects phishing accounts.

\begin{table}[t]
\centering
\caption{Dataset statistics.}
\label{tab:datasets}
\scriptsize
\resizebox{\columnwidth}{!}{%
\begin{tabular}{lrrrrr}
\toprule
Dataset & Nodes & Temporal edges & Minority & Majority & IR \\
\midrule
DGraph-Fin & 3,700,550 & 4,300,999 & 15,509 & 1,210,092 & 78.03 \\
Elliptic & 203,769 & 234,355 & 4,545 & 42,019 & 9.25 \\
Elliptic++ Transactions & 203,769 & 234,355 & 4,545 & 42,019 & 9.25 \\
Elliptic++ Actors & 822,942 & 2,868,964 & 14,266 & 251,088 & 17.60 \\
Ethereum & 2,973,489 & 13,551,303 & 1,165 & 2,972,324 & 2,551.35 \\
\bottomrule
\end{tabular}
}
\end{table}

\noindent\textbf{Compared Baselines.}
We compare with thirteen baselines from four categories: (1) temporal graph methods, TGC~\cite{liu2024tgc}, TGAT~\cite{xu2020tgat}, TGN~\cite{rossi2020tgn}, and DyGFormer~\cite{yu2023dygformer}; (2) static graph autoencoder methods, GraphPAE~\cite{liu2025graphpae} and DGMAE~\cite{zheng2025dgmae}; (3) imbalanced graph methods, GraphSMOTE~\cite{zhao2021graphsmote}, LTE4G~\cite{yun2022lte4g}, and GraphENS~\cite{park2022graphens}; and (4) graph diffusion methods, DDM~\cite{zhu2023ddm}, SDMG~\cite{zhu2025sdmg}, LGD~\cite{zhou2024lgd}, and TGAT+Conda~\cite{xu2020tgat,tian2024conda}.

\begin{table*}[t]
\centering
\caption{Node classification results (\%). The best and second-best results in each metric row are shown in bold and underlined, respectively.}
\label{tab:main-results}
\scriptsize
\setlength{\tabcolsep}{1.0pt}
\renewcommand{\arraystretch}{0.96}
\def\bestmath$#1${\ensuremath{\pmb{#1}}}
\providecommand{\best}[1]{\bestmath#1}
\resizebox{\textwidth}{!}{%
\begin{tabular}{@{}ll*{14}{c}@{}}
\toprule
\textbf{Dataset} & \textbf{Metric} & \textbf{TGC} & \textbf{DyGFormer} & \textbf{TGAT} & \textbf{TGN} & \textbf{GraphPAE} & \textbf{DGMAE} & \textbf{GraphSMOTE} & \textbf{LTE4G} & \textbf{GraphENS} & \textbf{DDM} & \textbf{SDMG} & \textbf{LGD} & \textbf{TGAT+Conda} & \textbf{\method} \\
\midrule
\multirow{5}{*}{DGraph-Fin} & AUROC & $72.89{\pm}.27$ & \underline{$76.10{\pm}.23$} & $72.06{\pm}.12$ & $73.41{\pm}.15$ & $70.92{\pm}.31$ & $70.61{\pm}.18$ & $72.46{\pm}.08$ & $71.90{\pm}.04$ & $71.85{\pm}.06$ & $72.34{\pm}.06$ & $74.59{\pm}.60$ & $75.48{\pm}.11$ & $74.38{\pm}.14$ & \best{$81.19{\pm}0.23$} \\
 & AUPRC & $2.89{\pm}.09$ & \underline{$3.69{\pm}.11$} & $3.14{\pm}.02$ & $3.12{\pm}.04$ & $2.92{\pm}.05$ & $3.00{\pm}.05$ & $2.73{\pm}.02$ & $2.64{\pm}.01$ & $3.07{\pm}.03$ & $3.32{\pm}.06$ & $3.54{\pm}.13$ & $3.67{\pm}.04$ & $3.25{\pm}.06$ & \best{$4.76{\pm}0.07$} \\
 & Recall-min. & $21.47{\pm}12.03$ & \underline{$32.01{\pm}6.46$} & $30.37{\pm}6.03$ & $28.95{\pm}7.42$ & $21.02{\pm}11.21$ & $18.30{\pm}4.44$ & $21.90{\pm}14.72$ & $20.68{\pm}11.92$ & $17.33{\pm}4.00$ & $19.25{\pm}5.21$ & $19.60{\pm}4.41$ & $25.16{\pm}7.80$ & $24.73{\pm}5.59$ & \best{$55.54{\pm}0.95$} \\
 & F1-min. & $5.64{\pm}.32$ & \underline{$7.91{\pm}.51$} & $6.82{\pm}.04$ & $6.61{\pm}.06$ & $5.90{\pm}.25$ & $6.75{\pm}.20$ & $5.50{\pm}.29$ & $5.45{\pm}.30$ & $6.34{\pm}.20$ & $6.95{\pm}.09$ & $7.18{\pm}.23$ & $7.47{\pm}.29$ & $6.64{\pm}.11$ & \best{$8.24{\pm}0.10$} \\
 & Macro-F1 & \underline{$49.69{\pm}0.02$} & $49.68{\pm}0.00$ & $49.68{\pm}0.00$ & $49.68{\pm}0.00$ & $49.68{\pm}0.00$ & $49.68{\pm}0.00$ & $49.68{\pm}0.00$ & $49.68{\pm}0.00$ & $49.68{\pm}0.00$ & $49.68{\pm}0.00$ & $49.68{\pm}0.00$ & $49.68{\pm}0.00$ & $49.68{\pm}0.00$ & \best{$49.72{\pm}0.06$} \\
\midrule
\multirow{5}{*}{Elliptic} & AUROC & \underline{$98.38{\pm}.13$} & $92.74{\pm}.44$ & $96.95{\pm}.13$ & \best{$98.47{\pm}.05$} & $97.07{\pm}.14$ & $94.88{\pm}.24$ & $95.53{\pm}.07$ & $96.54{\pm}.16$ & $98.28{\pm}.04$ & $97.94{\pm}.14$ & $98.29{\pm}.09$ & $95.23{\pm}.24$ & $98.33{\pm}.06$ & $98.21{\pm}0.06$ \\
 & AUPRC & \underline{$93.32{\pm}.49$} & $76.83{\pm}.65$ & $86.31{\pm}.68$ & $93.27{\pm}.23$ & $86.21{\pm}.60$ & $77.85{\pm}.39$ & $74.59{\pm}.55$ & $85.66{\pm}.39$ & $92.30{\pm}.07$ & $89.83{\pm}.41$ & $92.52{\pm}.26$ & $88.10{\pm}.29$ & $93.13{\pm}.38$ & \best{$93.61{\pm}0.15$} \\
 & Recall-min. & \underline{$85.23{\pm}1.54$} & $65.27{\pm}3.95$ & $76.59{\pm}4.07$ & $84.38{\pm}1.85$ & $73.69{\pm}4.09$ & $65.10{\pm}2.80$ & $68.88{\pm}3.54$ & $70.05{\pm}1.90$ & $82.11{\pm}1.36$ & $79.38{\pm}11.04$ & $85.02{\pm}3.49$ & $78.36{\pm}.81$ & $84.16{\pm}1.01$ & \best{$86.44{\pm}0.37$} \\
 & F1-min. & $87.82{\pm}.67$ & $70.63{\pm}1.95$ & $79.40{\pm}.96$ & \underline{$87.86{\pm}.59$} & $77.54{\pm}1.27$ & $69.92{\pm}.77$ & $69.68{\pm}1.16$ & $77.23{\pm}.81$ & $85.32{\pm}.20$ & $78.30{\pm}6.03$ & $85.03{\pm}6.47$ & $85.35{\pm}.19$ & $87.08{\pm}.91$ & \best{$88.54{\pm}0.31$} \\
 & Macro-F1 & $93.26{\pm}0.38$ & $75.65{\pm}4.48$ & $88.61{\pm}0.53$ & \underline{$93.36{\pm}0.29$} & $87.60{\pm}0.70$ & $83.42{\pm}0.38$ & $83.18{\pm}0.61$ & $87.73{\pm}0.47$ & $91.75{\pm}0.40$ & $87.90{\pm}3.46$ & $91.61{\pm}3.89$ & $91.93{\pm}0.10$ & $93.16{\pm}0.40$ & \best{$93.68{\pm}0.17$} \\
\midrule
\multirow{5}{*}{Elliptic++ Trans.} & AUROC & $98.07{\pm}0.01$ & $87.67{\pm}1.01$ & $96.31{\pm}.10$ & $98.07{\pm}.06$ & $96.99{\pm}.09$ & $95.28{\pm}.10$ & $94.61{\pm}.17$ & $96.39{\pm}.17$ & $97.90{\pm}.13$ & $97.33{\pm}.14$ & $95.33{\pm}.66$ & $92.36{\pm}.40$ & \underline{$98.08{\pm}0.03$} & \best{$98.32{\pm}0.04$} \\
 & AUPRC & $92.79{\pm}0.25$ & $68.47{\pm}.88$ & $85.04{\pm}.29$ & $92.28{\pm}.30$ & $85.35{\pm}.28$ & $64.83{\pm}.58$ & $72.88{\pm}.89$ & $85.25{\pm}.50$ & $90.31{\pm}.11$ & $89.60{\pm}.45$ & $84.67{\pm}1.69$ & $81.92{\pm}.94$ & \underline{$93.22{\pm}0.26$} & \best{$93.70{\pm}0.14$} \\
 & Recall-min. & $84.12{\pm}2.54$ & $56.16{\pm}4.75$ & $75.08{\pm}2.30$ & $84.20{\pm}1.80$ & $75.80{\pm}1.38$ & $41.48{\pm}6.18$ & $69.08{\pm}4.39$ & $74.37{\pm}1.94$ & \underline{$84.89{\pm}.39$} & $80.62{\pm}3.63$ & $68.37{\pm}3.26$ & $69.96{\pm}.75$ & $84.59{\pm}1.14$ & \best{$86.20{\pm}0.70$} \\
 & F1-min. & $87.31{\pm}0.23$ & $61.91{\pm}2.50$ & $79.18{\pm}.50$ & $87.58{\pm}.77$ & $77.43{\pm}.68$ & $52.69{\pm}3.92$ & $67.85{\pm}.92$ & $77.82{\pm}.72$ & $83.84{\pm}.32$ & $83.31{\pm}.94$ & $78.20{\pm}2.26$ & $78.30{\pm}.15$ & \underline{$87.89{\pm}0.59$} & \best{$88.70{\pm}0.29$} \\
 & Macro-F1 & \underline{$93.47{\pm}0.32$} & $54.82{\pm}3.10$ & $88.54{\pm}0.27$ & $93.22{\pm}0.34$ & $87.53{\pm}0.40$ & $82.01{\pm}0.84$ & $82.17{\pm}0.50$ & $87.74{\pm}0.47$ & $89.77{\pm}1.71$ & $90.79{\pm}0.54$ & $88.10{\pm}1.21$ & $88.13{\pm}0.08$ & $93.22{\pm}0.48$ & \best{$93.77{\pm}0.16$} \\
\midrule
\multirow{5}{*}{Elliptic++ Actors} & AUROC & \best{$98.85{\pm}0.06$} & $97.87{\pm}.09$ & $97.30{\pm}.09$ & $96.59{\pm}.13$ & $98.37{\pm}.07$ & $95.28{\pm}.10$ & $97.24{\pm}.10$ & $96.92{\pm}.10$ & \underline{$98.75{\pm}.04$} & $98.44{\pm}.08$ & $98.40{\pm}.08$ & $90.16{\pm}.83$ & $97.88{\pm}.06$ & $98.62{\pm}0.02$ \\
 & AUPRC & $89.69{\pm}0.82$ & $85.60{\pm}.42$ & $81.85{\pm}.53$ & $78.84{\pm}.73$ & $88.92{\pm}.42$ & $64.83{\pm}.58$ & $82.23{\pm}.38$ & $82.27{\pm}.53$ & \underline{$90.18{\pm}.33$} & $87.97{\pm}.47$ & $87.17{\pm}.48$ & $79.22{\pm}2.10$ & $84.66{\pm}.37$ & \best{$90.28{\pm}0.17$} \\
 & Recall-min. & $75.06{\pm}1.70$ & $68.55{\pm}1.93$ & $61.84{\pm}2.68$ & $60.62{\pm}4.30$ & $70.58{\pm}1.32$ & $41.48{\pm}6.18$ & $66.42{\pm}3.10$ & $67.86{\pm}2.24$ & \underline{$75.50{\pm}2.33$} & $72.35{\pm}3.31$ & $71.39{\pm}2.77$ & $65.49{\pm}3.45$ & $67.83{\pm}2.65$ & \best{$79.90{\pm}0.55$} \\
 & F1-min. & $81.21{\pm}0.97$ & $78.25{\pm}.43$ & $72.35{\pm}1.20$ & $70.38{\pm}1.99$ & $80.67{\pm}.68$ & $52.69{\pm}3.92$ & $74.78{\pm}1.37$ & $76.23{\pm}.65$ & \underline{$82.08{\pm}.53$} & $79.50{\pm}.70$ & $78.62{\pm}.89$ & $72.48{\pm}2.30$ & $76.73{\pm}.83$ & \best{$83.30{\pm}0.35$} \\
 & Macro-F1 & \underline{$90.75{\pm}0.48$} & $88.59{\pm}0.23$ & $85.52{\pm}0.61$ & $84.48{\pm}1.03$ & $89.86{\pm}0.39$ & $75.31{\pm}1.96$ & $86.76{\pm}0.74$ & $87.52{\pm}0.35$ & $90.58{\pm}0.28$ & $89.22{\pm}0.35$ & $88.77{\pm}0.46$ & $66.90{\pm}2.77$ & $87.73{\pm}0.64$ & \best{$91.33{\pm}0.26$} \\
\midrule
\multirow{5}{*}{Ethereum} & AUROC & $99.26{\pm}.09$ & $99.33{\pm}.17$ & \best{$99.63{\pm}.06$} & $99.29{\pm}.09$ & \underline{$99.48{\pm}.04$} & $97.88{\pm}.13$ & $99.22{\pm}.09$ & $99.24{\pm}.08$ & $99.28{\pm}.07$ & $98.43{\pm}.21$ & $99.15{\pm}.09$ & $91.01{\pm}.39$ & $99.17{\pm}.11$ & $99.18{\pm}0.09$ \\
 & AUPRC & $36.95{\pm}2.05$ & $46.35{\pm}.60$ & \underline{$47.40{\pm}2.69$} & $42.86{\pm}2.10$ & $42.94{\pm}2.11$ & $32.08{\pm}1.04$ & $38.65{\pm}1.42$ & $39.42{\pm}1.36$ & $40.76{\pm}1.52$ & $26.30{\pm}1.33$ & $33.09{\pm}2.55$ & $19.90{\pm}1.25$ & $34.34{\pm}1.92$ & \best{$50.07{\pm}0.89$} \\
 & Recall-min. & $23.24{\pm}12.98$ & $36.71{\pm}8.75$ & \underline{$39.61{\pm}11.50$} & $30.18{\pm}10.24$ & $34.91{\pm}4.40$ & $31.17{\pm}5.26$ & $21.85{\pm}6.93$ & $24.68{\pm}7.84$ & $27.95{\pm}7.26$ & $24.77{\pm}3.67$ & $11.89{\pm}15.29$ & $8.24{\pm}3.10$ & $23.61{\pm}9.33$ & \best{$54.55{\pm}2.99$} \\
 & F1-min. & $30.84{\pm}15.46$ & $46.08{\pm}7.65$ & \underline{$46.58{\pm}8.87$} & $38.64{\pm}9.87$ & $45.32{\pm}3.51$ & $40.28{\pm}3.54$ & $31.68{\pm}6.73$ & $34.21{\pm}7.18$ & $37.58{\pm}6.64$ & $31.79{\pm}1.03$ & $14.94{\pm}15.72$ & $14.96{\pm}4.20$ & $31.83{\pm}9.84$ & \best{$55.26{\pm}0.76$} \\
 & Macro-F1 & $66.82{\pm}6.26$ & $73.03{\pm}3.65$ & \underline{$74.99{\pm}3.16$} & $69.32{\pm}4.94$ & $66.64{\pm}11.61$ & $70.13{\pm}1.77$ & $65.83{\pm}3.55$ & $49.99{\pm}0.00$ & $68.79{\pm}3.32$ & $61.21{\pm}4.18$ & $57.46{\pm}7.86$ & $50.12{\pm}0.28$ & $65.91{\pm}4.92$ & \best{$77.60{\pm}0.39$} \\
\bottomrule
\end{tabular}%
}
\end{table*}

\noindent\textbf{Evaluation Protocol.}
All methods use the same data preprocessing, node splits, and downstream classifier. \method performs label-free self-supervised learning on the complete temporal graph, processing all edge events chronologically. DGraph-Fin follows its official 70\%/15\%/15\% split; labeled nodes in the other datasets are randomly divided into training, validation, and test sets at 60\%/20\%/20\% using a fixed seed. After representation learning, each encoder is frozen and evaluated using the same two-layer MLP.

We report AUROC, AUPRC, minority-class Recall, minority-class F1, and Macro-F1, treating the minority class as positive.

AUROC denotes the area under the receiver operating characteristic (ROC) curve and measures the model's ability to distinguish positive from negative samples across classification thresholds.

AUPRC summarizes the precision--recall trade-off across different classification thresholds and is computed as the area under the precision--recall curve. In its discrete form:
\begin{equation}
\mathrm{AUPRC}\approx\sum_{c=1}^{C}(R_c-R_{c-1})P_c,
\end{equation}
where $P_c$ and $R_c$ denote precision and recall at the $c$-th classification threshold, respectively. Minority-class Recall measures the proportion of minority-class nodes correctly identified, while minority-class F1 is the harmonic mean of minority-class precision and recall. Macro-F1 is the arithmetic mean of the class-wise F1 scores. Higher values indicate better performance for all metrics.


For DGraph-Fin, we search $[0.01,0.50]$ at intervals of $0.01$ and select the threshold maximizing validation minority-class F1; the other datasets use $0.5$. All results report the mean and standard deviation over ten random seeds.

\noindent\textbf{Implementation Details.}
Experiments run on an NVIDIA GeForce RTX 3090 GPU. For \method, we set the total number of diffusion steps to $Q=200$ and adopt a linear noise schedule. The conditional U-Net contains three encoder blocks and three decoder blocks. The low-frequency encoder uses a two-layer GAT with two attention heads and a hidden dimension of $384$. Baselines follow their official implementations, original training procedures, and recommended hyperparameters. The downstream classifier uses Adam~\cite{kingma2015adam} with a learning rate of $0.01$, weight decay of $5\times10^{-7}$, batch size $8{,}192$, and $200$ epochs. Hyperparameters are selected on the validation set.

Next, we seek to answer the following research questions:

\noindent\textbf{RQ1 (Classification Performance).}
How does \method compare with representative baselines for class-imbalanced node classification on temporal graphs?

\noindent\textbf{RQ2 (Ablation: Necessity of Components).}
Are the key components of \method necessary for its classification performance?

\noindent\textbf{RQ3 (Parameter Sensitivity).}
How sensitive is the performance of \method to changes in key hyperparameters?

\noindent\textbf{RQ4 (Core Mechanism Analysis).}
How do Distribution-Aware Selective Propagation and Multi-View Discriminative Fusion affect minority representation assimilation and class discriminability?

\subsection{Experimental Results}

\noindent\textbf{RQ1 (Classification Performance).}
As shown in Table~\ref{tab:main-results}, we observe that (i) compared with temporal graph methods (TGC, DyGFormer, TGAT, and TGN), \method achieves the best results on all non-AUROC metrics across the five datasets. On Ethereum, minority-class Recall and F1 improve by 14.94 and 8.68 percentage points, respectively, over the strongest temporal baselines. Its comparable AUROC further shows that explicitly accounting for class imbalance complements temporal modeling, particularly for minority-class recognition.

(ii) Compared with static graph autoencoder methods (GraphPAE and DGMAE), \method ranks first across all metrics on four datasets. On Ethereum, AUROC remains comparable, while minority-class Recall and F1 increase by 19.64 and 9.94 percentage points, respectively, reflecting the benefit of capturing temporal edge events and minority-related interaction patterns.

(iii) Compared with class-imbalanced graph methods (GraphENS, LTE4G, and GraphSMOTE), \method consistently achieves the best non-AUROC results. The advantage is especially clear on Ethereum, with gains of 26.60 and 17.68 percentage points in minority-class Recall and F1, respectively, while maintaining comparable AUROC. Such results reveal that class-imbalance handling can be further strengthened by incorporating temporal information.

(iv) For graph diffusion methods (DDM, SDMG, LGD, and TGAT+Conda), \method achieves the best results across all metrics on four datasets and trails only slightly in AUROC on Elliptic. On Ethereum, minority-class Recall and F1 are improved by 29.78 and 23.43 percentage points, respectively, showing the benefit of incorporating minority-relevant information into the diffusion condition.

Overall, \method consistently achieves stronger minority-class recognition across all four categories of baselines and maintains stable performance across datasets with varying imbalance levels, demonstrating its effectiveness and robustness.

\noindent\textbf{RQ2 (Ablation: Necessity of Components).}
Table~\ref{tab:ablation} shows that removing any core component degrades performance.

\begin{table*}[t]
\centering
\caption{Ablation results (\%). The best and second-best results in each column are shown in bold and underlined, respectively.}
\label{tab:ablation}
\scriptsize
\setlength{\tabcolsep}{1.2pt}
\renewcommand{\arraystretch}{1.05}
\begin{tabular*}{\textwidth}{@{\extracolsep{\fill}}l*{10}{c}@{}}
\toprule
\multirow{2}{*}{Variant}
& \multicolumn{5}{c}{DGraph-Fin}
& \multicolumn{5}{c}{Elliptic} \\
\cmidrule(lr){2-6}\cmidrule(lr){7-11}
& AUROC & AUPRC & Recall-min. & F1-min. & Macro-F1
& AUROC & AUPRC & Recall-min. & F1-min. & Macro-F1 \\
\midrule
w/o Temporal Edge-Event Construction
& 75.54$\pm$0.12 & 3.52$\pm$0.09 & 41.35$\pm$3.54 & 6.79$\pm$0.18 & 49.69$\pm$0.02
& 97.77$\pm$0.11 & 91.83$\pm$0.24 & 83.99$\pm$1.51 & 85.59$\pm$0.30 & 92.10$\pm$0.56 \\
w/o Distribution-Aware Selective Propagation
& 80.64$\pm$0.23 & 4.57$\pm$0.08 & 50.88$\pm$3.34 & \underline{8.11$\pm$0.07} & 49.68$\pm$0.00
& \underline{98.11$\pm$0.09} & \underline{93.13$\pm$0.13} & \underline{85.00$\pm$1.11} & \underline{86.71$\pm$0.48} & 92.85$\pm$0.56 \\
w/o Multi-View Discriminative Fusion
& 80.90$\pm$0.29 & 4.67$\pm$0.06 & \underline{53.44$\pm$0.99} & 8.08$\pm$0.04 & \underline{49.71$\pm$0.02}
& 97.93$\pm$0.18 & 92.35$\pm$0.50 & 83.96$\pm$1.92 & 86.29$\pm$0.41 & \underline{92.91$\pm$0.39} \\
w/o Debiased Contrastive Learning
& \underline{81.01$\pm$0.39} & \underline{4.71$\pm$0.19} & 50.07$\pm$7.76 & 8.09$\pm$0.05 & 49.69$\pm$0.02
& 97.75$\pm$0.14 & 91.63$\pm$0.18 & 83.34$\pm$1.25 & 86.04$\pm$0.66 & 92.31$\pm$0.38 \\
\midrule
Full Model
& \textbf{81.19$\pm$0.23} & \textbf{4.76$\pm$0.07} & \textbf{55.54$\pm$0.95} & \textbf{8.24$\pm$0.10} & \textbf{49.72$\pm$0.06}
& \textbf{98.21$\pm$0.06} & \textbf{93.61$\pm$0.15} & \textbf{86.44$\pm$0.37} & \textbf{88.54$\pm$0.31} & \textbf{93.68$\pm$0.17} \\
\bottomrule
\end{tabular*}
\end{table*}

\begin{figure*}[t]
\centering
\begin{minipage}[t]{0.48\textwidth}
\centering
\includegraphics[width=0.94\linewidth]{figures/hyperparameter_sensitivity.png}
\captionof{figure}{Hyperparameter sensitivity of $\tau$, $\tau_{+}$, $k$, and $\tau_c$}
\label{fig:sensitivity}
\end{minipage}
\hfill
\begin{minipage}[t]{0.48\textwidth}
\centering
\includegraphics[width=0.86\linewidth]{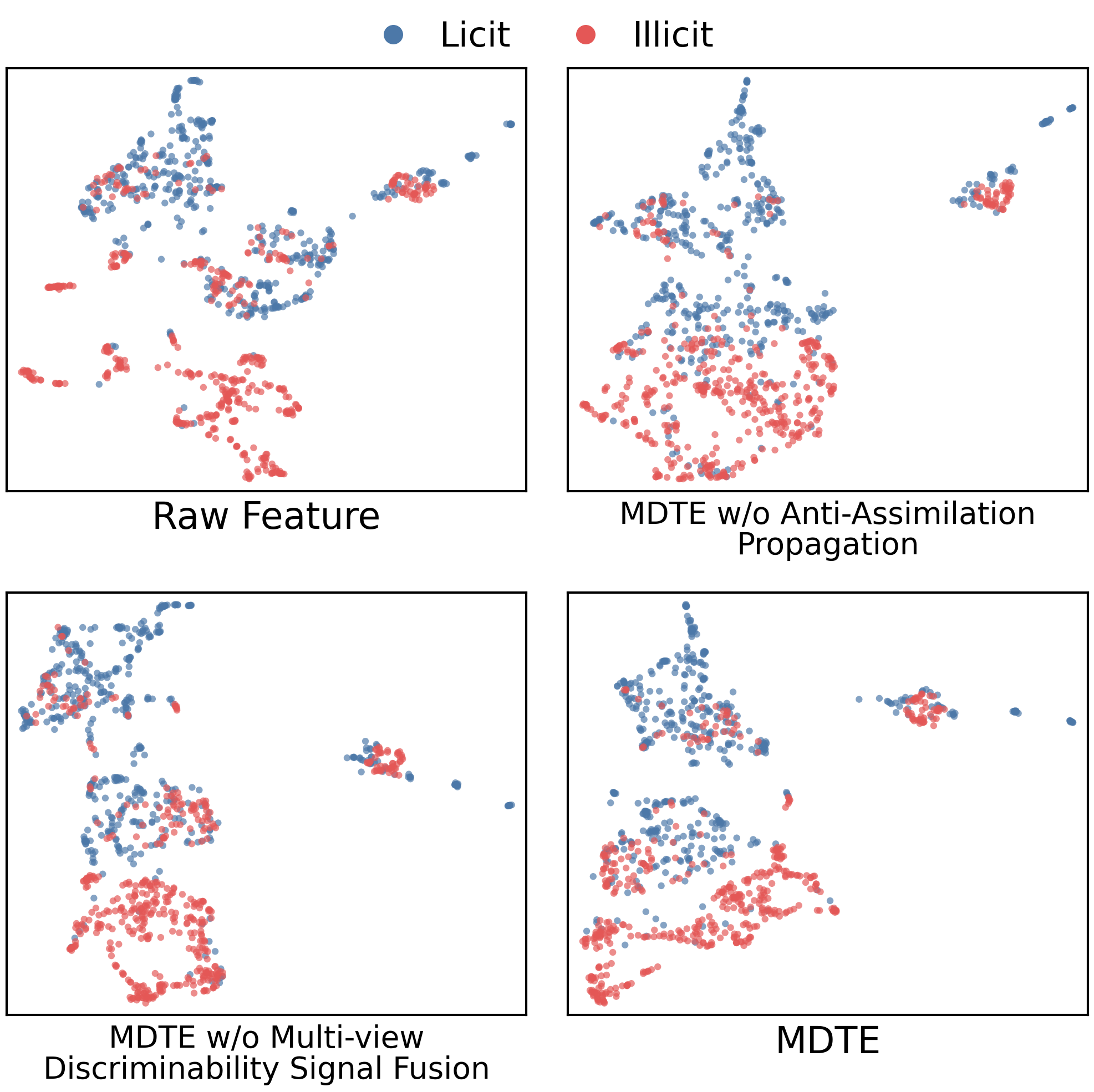}
\captionof{figure}{Visualization of Node Representations on Elliptic} 
\label{fig:umap}
\end{minipage}
\end{figure*}

Removing Temporal Edge-Event Construction reduces minority-class Recall by 14.19 percentage points on DGraph-Fin. Without Distribution-Aware Selective Propagation, minority-class Recall decreases by 4.66 percentage points on DGraph-Fin. Without Multi-View Discriminative Fusion, minority-class Recall decreases by 2.48 percentage points on Elliptic. Without Debiased Contrastive Learning, minority-class Recall drops by 5.47 percentage points on DGraph-Fin. These results confirm that Temporal Edge-Event Construction captures temporal interactions; Distribution-Aware Selective Propagation suppresses propagation interference and majority-class assimilation while retaining useful dependencies; Multi-View Discriminative Fusion enhances discriminative evidence for minority representations; and Debiased Contrastive Learning maintains consistency between original node representations and denoised interaction-aggregated representations.

\noindent\textbf{RQ3 (Parameter Sensitivity).}
Figure~\ref{fig:sensitivity} evaluates four hyperparameters: the propagation-filtering threshold $\tau$ for the node mask $\mathbf{m}$ in Eq.~\eqref{eq:filtered-adjacency}, the LOF neighborhood size $k$ in Eq.~\eqref{eq:reachability-distance}, and the potential-false-negative prior $\tau_{+}$ and contrastive temperature $\tau_c$ in Eq.~\eqref{eq:debiased-negative}.
Mean-centered F1-min is calculated by subtracting the mean minority-class F1 across all tested settings from the minority-class F1 at each setting. The settings $\tau=0.85$, $\tau_{+}=0.025$, $k=30$, and $\tau_c=0.05$ yield favorable performance across datasets. Performance remains relatively stable on four datasets, demonstrating good robustness to parameter variations. Ethereum is more sensitive due to its extremely high class imbalance and limited minority samples, making minority-class performance more susceptible to parameter changes.

\noindent\textbf{RQ4 (Core Mechanism Analysis).}
Figure~\ref{fig:umap} uses UMAP~\cite{mcinnes2018umap}to visualize 500 licit and 500 illicit nodes from the Elliptic test set. Removing Distribution-Aware Selective Propagation causes illicit nodes to move toward dense licit regions, indicating increased minority representation assimilation. Without Multi-View Discriminative Fusion, the class boundary becomes less distinct, reflecting weaker discriminability for minority representations. In contrast, the full model produces less overlap and clearer class boundaries, validating the effectiveness of both mechanisms.


\section{Conclusion}
This paper proposes \method, a minority-aware diffusion framework over temporal edge events for imbalanced node classification, to accurately identify minority-class nodes. \method develops two complementary mechanisms, Distribution-Aware Selective Propagation and Multi-View Discriminative Fusion, to mitigate majority-class interference and enhance minority-class discriminability. Extensive experiments on five real-world datasets validate the effectiveness of \method. In future work, we plan to further improve the discriminability and robustness of minority-node representations.

\section{Ethical Considerations}
This work studies node classification on imbalanced temporal graphs, with
potential applications to fraud, phishing, and illicit-activity detection.
False positives cause legitimate accounts to be misclassified. Therefore,
\method should support rather than replace human review in consequential
decisions. Practical deployments should calibrate classification thresholds to
application-specific costs and evaluate performance across relevant subgroups.
All experiments use publicly released research datasets.

\FloatBarrier

\bibliographystyle{ACM-Reference-Format}
\bibliography{references}


\begin{thebibliography}{45}


\ifx \showCODEN    \undefined \def \showCODEN     #1{\unskip}     \fi
\ifx \showISBNx    \undefined \def \showISBNx     #1{\unskip}     \fi
\ifx \showISBNxiii \undefined \def \showISBNxiii  #1{\unskip}     \fi
\ifx \showISSN     \undefined \def \showISSN      #1{\unskip}     \fi
\ifx \showLCCN     \undefined \def \showLCCN      #1{\unskip}     \fi
\ifx \shownote     \undefined \def \shownote      #1{#1}          \fi
\ifx \showarticletitle \undefined \def \showarticletitle #1{#1}   \fi
\ifx \showURL      \undefined \def \showURL       {\relax}        \fi
\providecommand\bibfield[2]{#2}
\providecommand\bibinfo[2]{#2}
\providecommand\natexlab[1]{#1}
\providecommand\showeprint[2][]{arXiv:#2}

\bibitem[Amini et~al\mbox{.}(2020)]%
        {amini2020deep}
\bibfield{author}{\bibinfo{person}{Alexander Amini}, \bibinfo{person}{Wilko Schwarting}, \bibinfo{person}{Ava Soleimany}, {and} \bibinfo{person}{Daniela Rus}.} \bibinfo{year}{2020}\natexlab{}.
\newblock \showarticletitle{Deep Evidential Regression}. In \bibinfo{booktitle}{\emph{Advances in Neural Information Processing Systems}}, Vol.~\bibinfo{volume}{33}.
\newblock
\urldef\tempurl%
\url{https://proceedings.neurips.cc/paper/2020/hash/aab085461de182608ee9f607f3f7d18f-Abstract.html}
\showURL{%
\tempurl}


\bibitem[Belkin et~al\mbox{.}(2006)]%
        {belkin2006manifold}
\bibfield{author}{\bibinfo{person}{Mikhail Belkin}, \bibinfo{person}{Partha Niyogi}, {and} \bibinfo{person}{Vikas Sindhwani}.} \bibinfo{year}{2006}\natexlab{}.
\newblock \showarticletitle{Manifold Regularization: A Geometric Framework for Learning from Labeled and Unlabeled Examples}.
\newblock \bibinfo{journal}{\emph{Journal of Machine Learning Research}}  \bibinfo{volume}{7} (\bibinfo{year}{2006}), \bibinfo{pages}{2399--2434}.
\newblock
\urldef\tempurl%
\url{https://www.jmlr.org/papers/v7/belkin06a.html}
\showURL{%
\tempurl}


\bibitem[Breunig et~al\mbox{.}(2000)]%
        {breunig2000lof}
\bibfield{author}{\bibinfo{person}{Markus~M. Breunig}, \bibinfo{person}{Hans-Peter Kriegel}, \bibinfo{person}{Raymond~T. Ng}, {and} \bibinfo{person}{J{\"o}rg Sander}.} \bibinfo{year}{2000}\natexlab{}.
\newblock \showarticletitle{{LOF}: Identifying Density-Based Local Outliers}. In \bibinfo{booktitle}{\emph{Proceedings of the 2000 ACM SIGMOD International Conference on Management of Data}}. \bibinfo{pages}{93--104}.
\newblock
\href{https://doi.org/10.1145/335191.335388}{doi:\nolinkurl{10.1145/335191.335388}}


\bibitem[Chen et~al\mbox{.}(2021a)]%
        {chen2021renode}
\bibfield{author}{\bibinfo{person}{Deli Chen}, \bibinfo{person}{Yankai Lin}, \bibinfo{person}{Guangxiang Zhao}, \bibinfo{person}{Xuancheng Ren}, \bibinfo{person}{Peng Li}, \bibinfo{person}{Jie Zhou}, {and} \bibinfo{person}{Xu Sun}.} \bibinfo{year}{2021}\natexlab{a}.
\newblock \showarticletitle{Topology-Imbalance Learning for Semi-Supervised Node Classification}. In \bibinfo{booktitle}{\emph{Advances in Neural Information Processing Systems}}, Vol.~\bibinfo{volume}{34}. \bibinfo{pages}{29885--29897}.
\newblock


\bibitem[Chen et~al\mbox{.}(2025)]%
        {chen2025graffe}
\bibfield{author}{\bibinfo{person}{Dingshuo Chen}, \bibinfo{person}{Shuchen Xue}, \bibinfo{person}{Liuji Chen}, \bibinfo{person}{Yingheng Wang}, \bibinfo{person}{Qiang Liu}, \bibinfo{person}{Shu Wu}, \bibinfo{person}{Zhi-Ming Ma}, {and} \bibinfo{person}{Liang Wang}.} \bibinfo{year}{2025}\natexlab{}.
\newblock \showarticletitle{{Graffe}: Graph Representation Learning via Diffusion Probabilistic Models}.
\newblock \bibinfo{journal}{\emph{arXiv preprint arXiv:2505.04956}} (\bibinfo{year}{2025}).
\newblock
\urldef\tempurl%
\url{https://arxiv.org/abs/2505.04956}
\showURL{%
\tempurl}


\bibitem[Chen et~al\mbox{.}(2021b)]%
        {chen2021phishing}
\bibfield{author}{\bibinfo{person}{Liang Chen}, \bibinfo{person}{Jiaying Peng}, \bibinfo{person}{Yang Liu}, \bibinfo{person}{Jintang Li}, \bibinfo{person}{Fenfang Xie}, {and} \bibinfo{person}{Zibin Zheng}.} \bibinfo{year}{2021}\natexlab{b}.
\newblock \showarticletitle{Phishing Scams Detection in {Ethereum} Transaction Network}.
\newblock \bibinfo{journal}{\emph{ACM Transactions on Internet Technology}} \bibinfo{volume}{21}, \bibinfo{number}{1} (\bibinfo{year}{2021}), \bibinfo{pages}{10:1--10:16}.
\newblock
\href{https://doi.org/10.1145/3398071}{doi:\nolinkurl{10.1145/3398071}}


\bibitem[Chuang et~al\mbox{.}(2020)]%
        {chuang2020debiased}
\bibfield{author}{\bibinfo{person}{Ching-Yao Chuang}, \bibinfo{person}{Joshua Robinson}, \bibinfo{person}{Yen-Chen Lin}, \bibinfo{person}{Antonio Torralba}, {and} \bibinfo{person}{Stefanie Jegelka}.} \bibinfo{year}{2020}\natexlab{}.
\newblock \showarticletitle{Debiased Contrastive Learning}. In \bibinfo{booktitle}{\emph{Advances in Neural Information Processing Systems}}, Vol.~\bibinfo{volume}{33}.
\newblock
\urldef\tempurl%
\url{https://proceedings.neurips.cc/paper/2020/hash/63c3ddcc7b23daa1e42dc41f9a44a873-Abstract.html}
\showURL{%
\tempurl}


\bibitem[Dai et~al\mbox{.}(2024)]%
        {dai2024diffgcl}
\bibfield{author}{\bibinfo{person}{Qi Dai}, \bibinfo{person}{Yumeng Song}, \bibinfo{person}{Yu Gu}, \bibinfo{person}{Fangfang Li}, {and} \bibinfo{person}{Xiaohua Li}.} \bibinfo{year}{2024}\natexlab{}.
\newblock \showarticletitle{Diffusion Model-Enhanced Contrastive Learning for Graph Representation}. In \bibinfo{booktitle}{\emph{Database Systems for Advanced Applications}} \emph{(\bibinfo{series}{Lecture Notes in Computer Science}, Vol.~\bibinfo{volume}{14855})}. \bibinfo{pages}{332--341}.
\newblock
\href{https://doi.org/10.1007/978-981-97-5572-1_22}{doi:\nolinkurl{10.1007/978-981-97-5572-1_22}}


\bibitem[Duan et~al\mbox{.}(2022)]%
        {duan2022sorag}
\bibfield{author}{\bibinfo{person}{Yijun Duan}, \bibinfo{person}{Xin Liu}, \bibinfo{person}{Adam Jatowt}, \bibinfo{person}{Hai~Tao Yu}, \bibinfo{person}{Steven Lynden}, \bibinfo{person}{Kyoung-Sook Kim}, {and} \bibinfo{person}{Akiyoshi Matono}.} \bibinfo{year}{2022}\natexlab{}.
\newblock \showarticletitle{{SORAG}: Synthetic Data Over-Sampling Strategy on Multi-Label Graphs}.
\newblock \bibinfo{journal}{\emph{Remote Sensing}} \bibinfo{volume}{14}, \bibinfo{number}{18} (\bibinfo{year}{2022}), \bibinfo{pages}{4479}.
\newblock
\href{https://doi.org/10.3390/rs14184479}{doi:\nolinkurl{10.3390/rs14184479}}


\bibitem[Duc and Ta(2026)]%
        {duc2026sdg}
\bibfield{author}{\bibinfo{person}{Nguyen~Minh Duc} {and} \bibinfo{person}{Viet~Cuong Ta}.} \bibinfo{year}{2026}\natexlab{}.
\newblock \showarticletitle{Sequence Diffusion Model for Temporal Link Prediction in Continuous-Time Dynamic Graph}.
\newblock \bibinfo{journal}{\emph{arXiv preprint arXiv:2601.23233}} (\bibinfo{year}{2026}).
\newblock
\urldef\tempurl%
\url{https://arxiv.org/abs/2601.23233}
\showURL{%
\tempurl}


\bibitem[Elmougy and Liu(2023)]%
        {elmougy2023ellipticpp}
\bibfield{author}{\bibinfo{person}{Youssef Elmougy} {and} \bibinfo{person}{Ling Liu}.} \bibinfo{year}{2023}\natexlab{}.
\newblock \showarticletitle{Demystifying Fraudulent Transactions and Illicit Nodes in the {Bitcoin} Network for Financial Forensics}. In \bibinfo{booktitle}{\emph{Proceedings of the 29th ACM SIGKDD Conference on Knowledge Discovery and Data Mining}}. \bibinfo{pages}{3979--3990}.
\newblock
\href{https://doi.org/10.1145/3580305.3599803}{doi:\nolinkurl{10.1145/3580305.3599803}}


\bibitem[Ho et~al\mbox{.}(2020)]%
        {ho2020denoising}
\bibfield{author}{\bibinfo{person}{Jonathan Ho}, \bibinfo{person}{Ajay Jain}, {and} \bibinfo{person}{Pieter Abbeel}.} \bibinfo{year}{2020}\natexlab{}.
\newblock \showarticletitle{Denoising Diffusion Probabilistic Models}. In \bibinfo{booktitle}{\emph{Advances in Neural Information Processing Systems}}, Vol.~\bibinfo{volume}{33}.
\newblock
\urldef\tempurl%
\url{https://proceedings.neurips.cc/paper/2020/hash/4c5bcfec8584af0d967f1ab10179ca4b-Abstract.html}
\showURL{%
\tempurl}


\bibitem[Huang et~al\mbox{.}(2022)]%
        {huang2022dgraph}
\bibfield{author}{\bibinfo{person}{Xuanwen Huang}, \bibinfo{person}{Yang Yang}, \bibinfo{person}{Yang Wang}, \bibinfo{person}{Chunping Wang}, \bibinfo{person}{Zhisheng Zhang}, \bibinfo{person}{Jiarong Xu}, \bibinfo{person}{Lei Chen}, {and} \bibinfo{person}{Michalis Vazirgiannis}.} \bibinfo{year}{2022}\natexlab{}.
\newblock \showarticletitle{{DGraph}: A Large-Scale Financial Dataset for Graph Anomaly Detection}. In \bibinfo{booktitle}{\emph{Advances in Neural Information Processing Systems}}, Vol.~\bibinfo{volume}{35}.
\newblock
\urldef\tempurl%
\url{https://proceedings.neurips.cc/paper_files/paper/2022/hash/8f1918f71972789db39ec0d85bb31110-Abstract-Datasets_and_Benchmarks.html}
\showURL{%
\tempurl}


\bibitem[Kingma and Ba(2015)]%
        {kingma2015adam}
\bibfield{author}{\bibinfo{person}{Diederik~P. Kingma} {and} \bibinfo{person}{Jimmy Ba}.} \bibinfo{year}{2015}\natexlab{}.
\newblock \showarticletitle{Adam: A Method for Stochastic Optimization}. In \bibinfo{booktitle}{\emph{International Conference on Learning Representations}}.
\newblock
\urldef\tempurl%
\url{https://arxiv.org/abs/1412.6980}
\showURL{%
\tempurl}


\bibitem[Liu et~al\mbox{.}(2024)]%
        {liu2024tgc}
\bibfield{author}{\bibinfo{person}{Meng Liu}, \bibinfo{person}{Yue Liu}, \bibinfo{person}{Ke Liang}, \bibinfo{person}{Wenxuan Tu}, \bibinfo{person}{Siwei Wang}, \bibinfo{person}{Sihang Zhou}, {and} \bibinfo{person}{Xinwang Liu}.} \bibinfo{year}{2024}\natexlab{}.
\newblock \showarticletitle{Deep Temporal Graph Clustering}. In \bibinfo{booktitle}{\emph{International Conference on Learning Representations}}.
\newblock
\urldef\tempurl%
\url{https://openreview.net/forum?id=ViNe1fjGME}
\showURL{%
\tempurl}


\bibitem[Liu et~al\mbox{.}(2025)]%
        {liu2025graphpae}
\bibfield{author}{\bibinfo{person}{Yang Liu}, \bibinfo{person}{Deyu Bo}, \bibinfo{person}{Wenxuan Cao}, \bibinfo{person}{Yuan Fang}, \bibinfo{person}{Yawen Li}, {and} \bibinfo{person}{Chuan Shi}.} \bibinfo{year}{2025}\natexlab{}.
\newblock \showarticletitle{Graph Positional Autoencoders as Self-supervised Learners}. In \bibinfo{booktitle}{\emph{Proceedings of the 31st ACM SIGKDD Conference on Knowledge Discovery and Data Mining}}. \bibinfo{pages}{1867--1878}.
\newblock
\href{https://doi.org/10.1145/3711896.3736990}{doi:\nolinkurl{10.1145/3711896.3736990}}


\bibitem[Liu et~al\mbox{.}(2022)]%
        {liu2022gatsmote}
\bibfield{author}{\bibinfo{person}{Yongxu Liu}, \bibinfo{person}{Zhi Zhang}, \bibinfo{person}{Yan Liu}, {and} \bibinfo{person}{Yao Zhu}.} \bibinfo{year}{2022}\natexlab{}.
\newblock \showarticletitle{{GATSMOTE}: Improving Imbalanced Node Classification on Graphs via Attention and Homophily}.
\newblock \bibinfo{journal}{\emph{Mathematics}} \bibinfo{volume}{10}, \bibinfo{number}{11} (\bibinfo{year}{2022}), \bibinfo{pages}{1799}.
\newblock
\href{https://doi.org/10.3390/math10111799}{doi:\nolinkurl{10.3390/math10111799}}


\bibitem[Lloyd(1982)]%
        {lloyd1982least}
\bibfield{author}{\bibinfo{person}{Stuart~P. Lloyd}.} \bibinfo{year}{1982}\natexlab{}.
\newblock \showarticletitle{Least Squares Quantization in {PCM}}.
\newblock \bibinfo{journal}{\emph{IEEE Transactions on Information Theory}} \bibinfo{volume}{28}, \bibinfo{number}{2} (\bibinfo{year}{1982}), \bibinfo{pages}{129--137}.
\newblock
\href{https://doi.org/10.1109/TIT.1982.1056489}{doi:\nolinkurl{10.1109/TIT.1982.1056489}}


\bibitem[Ma et~al\mbox{.}(2025)]%
        {ma2025cilg}
\bibfield{author}{\bibinfo{person}{Yihong Ma}, \bibinfo{person}{Yijun Tian}, \bibinfo{person}{Nuno Moniz}, {and} \bibinfo{person}{Nitesh~V. Chawla}.} \bibinfo{year}{2025}\natexlab{}.
\newblock \showarticletitle{Class-Imbalanced Learning on Graphs: A Survey}.
\newblock \bibinfo{journal}{\emph{Comput. Surveys}} \bibinfo{volume}{57}, \bibinfo{number}{8} (\bibinfo{year}{2025}), \bibinfo{pages}{1--16}.
\newblock
\href{https://doi.org/10.1145/3718734}{doi:\nolinkurl{10.1145/3718734}}


\bibitem[McInnes et~al\mbox{.}(2018)]%
        {mcinnes2018umap}
\bibfield{author}{\bibinfo{person}{Leland McInnes}, \bibinfo{person}{John Healy}, \bibinfo{person}{Nathaniel Saul}, {and} \bibinfo{person}{Lukas Grossberger}.} \bibinfo{year}{2018}\natexlab{}.
\newblock \showarticletitle{{UMAP}: Uniform Manifold Approximation and Projection}.
\newblock \bibinfo{journal}{\emph{Journal of Open Source Software}} \bibinfo{volume}{3}, \bibinfo{number}{29} (\bibinfo{year}{2018}), \bibinfo{pages}{861}.
\newblock
\href{https://doi.org/10.21105/joss.00861}{doi:\nolinkurl{10.21105/joss.00861}}


\bibitem[Park et~al\mbox{.}(2022)]%
        {park2022graphens}
\bibfield{author}{\bibinfo{person}{Joonhyung Park}, \bibinfo{person}{Jaeyun Song}, {and} \bibinfo{person}{Eunho Yang}.} \bibinfo{year}{2022}\natexlab{}.
\newblock \showarticletitle{{GraphENS}: Neighbor-Aware Ego Network Synthesis for Class-Imbalanced Node Classification}. In \bibinfo{booktitle}{\emph{International Conference on Learning Representations}}.
\newblock
\urldef\tempurl%
\url{https://openreview.net/forum?id=MXEl7i-iru}
\showURL{%
\tempurl}


\bibitem[Qu et~al\mbox{.}(2021)]%
        {qu2021imgagn}
\bibfield{author}{\bibinfo{person}{Liang Qu}, \bibinfo{person}{Huaisheng Zhu}, \bibinfo{person}{Ruiqi Zheng}, \bibinfo{person}{Yuhui Shi}, {and} \bibinfo{person}{Hongzhi Yin}.} \bibinfo{year}{2021}\natexlab{}.
\newblock \showarticletitle{{ImGAGN}: Imbalanced Network Embedding via Generative Adversarial Graph Networks}. In \bibinfo{booktitle}{\emph{Proceedings of the 27th ACM SIGKDD Conference on Knowledge Discovery and Data Mining}}. \bibinfo{pages}{1390--1398}.
\newblock
\href{https://doi.org/10.1145/3447548.3467334}{doi:\nolinkurl{10.1145/3447548.3467334}}


\bibitem[Ramp{\v{s}}ek et~al\mbox{.}(2022)]%
        {rampasek2022recipe}
\bibfield{author}{\bibinfo{person}{Ladislav Ramp{\v{s}}ek}, \bibinfo{person}{Michael Galkin}, \bibinfo{person}{Vijay~Prakash Dwivedi}, \bibinfo{person}{Anh~Tuan Luu}, \bibinfo{person}{Guy Wolf}, {and} \bibinfo{person}{Dominique Beaini}.} \bibinfo{year}{2022}\natexlab{}.
\newblock \showarticletitle{Recipe for a General, Powerful, Scalable Graph Transformer}. In \bibinfo{booktitle}{\emph{Advances in Neural Information Processing Systems}}, Vol.~\bibinfo{volume}{35}.
\newblock
\urldef\tempurl%
\url{https://proceedings.neurips.cc/paper_files/paper/2022/hash/5d4834a159f1547b267a05a4e2b7cf5e-Abstract-Conference.html}
\showURL{%
\tempurl}


\bibitem[Rossi et~al\mbox{.}(2020)]%
        {rossi2020tgn}
\bibfield{author}{\bibinfo{person}{Emanuele Rossi}, \bibinfo{person}{Ben Chamberlain}, \bibinfo{person}{Fabrizio Frasca}, \bibinfo{person}{Davide Eynard}, \bibinfo{person}{Federico Monti}, {and} \bibinfo{person}{Michael~M. Bronstein}.} \bibinfo{year}{2020}\natexlab{}.
\newblock \showarticletitle{Temporal Graph Networks for Deep Learning on Dynamic Graphs}.
\newblock \bibinfo{journal}{\emph{arXiv preprint arXiv:2006.10637}} (\bibinfo{year}{2020}).
\newblock
\urldef\tempurl%
\url{https://arxiv.org/abs/2006.10637}
\showURL{%
\tempurl}


\bibitem[Shi et~al\mbox{.}(2020)]%
        {shi2020drgcn}
\bibfield{author}{\bibinfo{person}{Min Shi}, \bibinfo{person}{Yufei Tang}, \bibinfo{person}{Xingquan Zhu}, \bibinfo{person}{David~A. Wilson}, {and} \bibinfo{person}{Jianxun Liu}.} \bibinfo{year}{2020}\natexlab{}.
\newblock \showarticletitle{Multi-Class Imbalanced Graph Convolutional Network Learning}. In \bibinfo{booktitle}{\emph{Proceedings of the Twenty-Ninth International Joint Conference on Artificial Intelligence}}. \bibinfo{pages}{2879--2885}.
\newblock
\href{https://doi.org/10.24963/ijcai.2020/398}{doi:\nolinkurl{10.24963/ijcai.2020/398}}


\bibitem[Song et~al\mbox{.}(2022)]%
        {song2022tam}
\bibfield{author}{\bibinfo{person}{Jaeyun Song}, \bibinfo{person}{Joonhyung Park}, {and} \bibinfo{person}{Eunho Yang}.} \bibinfo{year}{2022}\natexlab{}.
\newblock \showarticletitle{{TAM}: Topology-Aware Margin Loss for Class-Imbalanced Node Classification}. In \bibinfo{booktitle}{\emph{Proceedings of the 39th International Conference on Machine Learning}} \emph{(\bibinfo{series}{Proceedings of Machine Learning Research}, Vol.~\bibinfo{volume}{162})}. \bibinfo{pages}{20369--20383}.
\newblock
\urldef\tempurl%
\url{https://proceedings.mlr.press/v162/song22a.html}
\showURL{%
\tempurl}


\bibitem[Thorndike(1953)]%
        {thorndike1953family}
\bibfield{author}{\bibinfo{person}{Robert~L. Thorndike}.} \bibinfo{year}{1953}\natexlab{}.
\newblock \showarticletitle{Who Belongs in the Family?}
\newblock \bibinfo{journal}{\emph{Psychometrika}} \bibinfo{volume}{18}, \bibinfo{number}{4} (\bibinfo{year}{1953}), \bibinfo{pages}{267--276}.
\newblock
\href{https://doi.org/10.1007/BF02289263}{doi:\nolinkurl{10.1007/BF02289263}}


\bibitem[Tian et~al\mbox{.}(2024)]%
        {tian2024conda}
\bibfield{author}{\bibinfo{person}{Yuxing Tian}, \bibinfo{person}{Yiyan Qi}, \bibinfo{person}{Aiwen Jiang}, \bibinfo{person}{Qi Huang}, {and} \bibinfo{person}{Jian Guo}.} \bibinfo{year}{2024}\natexlab{}.
\newblock \showarticletitle{Latent Conditional Diffusion-based Data Augmentation for Continuous-Time Dynamic Graph Model}.
\newblock \bibinfo{journal}{\emph{arXiv preprint arXiv:2407.08500}} (\bibinfo{year}{2024}).
\newblock
\urldef\tempurl%
\url{https://arxiv.org/abs/2407.08500}
\showURL{%
\tempurl}


\bibitem[Vignac et~al\mbox{.}(2023)]%
        {vignac2023digress}
\bibfield{author}{\bibinfo{person}{Cl{\'e}ment Vignac}, \bibinfo{person}{Igor Krawczuk}, \bibinfo{person}{Antoine Siraudin}, \bibinfo{person}{Bohan Wang}, \bibinfo{person}{Volkan Cevher}, {and} \bibinfo{person}{Pascal Frossard}.} \bibinfo{year}{2023}\natexlab{}.
\newblock \showarticletitle{{DiGress}: Discrete Denoising Diffusion for Graph Generation}. In \bibinfo{booktitle}{\emph{International Conference on Learning Representations}}.
\newblock
\urldef\tempurl%
\url{https://openreview.net/forum?id=UaAD-Nu86WX}
\showURL{%
\tempurl}


\bibitem[Wang et~al\mbox{.}(2022)]%
        {wang2022egcn}
\bibfield{author}{\bibinfo{person}{Kefan Wang}, \bibinfo{person}{Jing An}, {and} \bibinfo{person}{Qi Kang}.} \bibinfo{year}{2022}\natexlab{}.
\newblock \showarticletitle{Effective-Aggregation Graph Convolutional Network for Imbalanced Classification}. In \bibinfo{booktitle}{\emph{2022 IEEE International Conference on Networking, Sensing and Control}}. \bibinfo{pages}{1--5}.
\newblock
\href{https://doi.org/10.1109/ICNSC55942.2022.10004069}{doi:\nolinkurl{10.1109/ICNSC55942.2022.10004069}}


\bibitem[Wang et~al\mbox{.}(2018)]%
        {wang2018rsdne}
\bibfield{author}{\bibinfo{person}{Zheng Wang}, \bibinfo{person}{Xiaojun Ye}, \bibinfo{person}{Chaokun Wang}, \bibinfo{person}{Yuexin Wu}, \bibinfo{person}{Changping Wang}, {and} \bibinfo{person}{Kaiwen Liang}.} \bibinfo{year}{2018}\natexlab{}.
\newblock \showarticletitle{{RSDNE}: Exploring Relaxed Similarity and Dissimilarity from Completely-Imbalanced Labels for Network Embedding}. In \bibinfo{booktitle}{\emph{Proceedings of the AAAI Conference on Artificial Intelligence}}, Vol.~\bibinfo{volume}{32}.
\newblock
\href{https://doi.org/10.1609/aaai.v32i1.11242}{doi:\nolinkurl{10.1609/aaai.v32i1.11242}}


\bibitem[Weber et~al\mbox{.}(2019)]%
        {weber2019elliptic}
\bibfield{author}{\bibinfo{person}{Mark Weber}, \bibinfo{person}{Giacomo Domeniconi}, \bibinfo{person}{Jie Chen}, \bibinfo{person}{Daniel Karl~I. Weidele}, \bibinfo{person}{Claudio Bellei}, \bibinfo{person}{Tom Robinson}, {and} \bibinfo{person}{Charles~E. Leiserson}.} \bibinfo{year}{2019}\natexlab{}.
\newblock \showarticletitle{Anti-Money Laundering in {Bitcoin}: Experimenting with Graph Convolutional Networks for Financial Forensics}.
\newblock \bibinfo{journal}{\emph{arXiv preprint arXiv:1908.02591}} (\bibinfo{year}{2019}).
\newblock
\urldef\tempurl%
\url{https://arxiv.org/abs/1908.02591}
\showURL{%
\tempurl}


\bibitem[Wei et~al\mbox{.}(2025)]%
        {wei2025gel}
\bibfield{author}{\bibinfo{person}{Chunyu Wei}, \bibinfo{person}{Wenji Hu}, \bibinfo{person}{Xingjia Hao}, \bibinfo{person}{Yunhai Wang}, \bibinfo{person}{Yueguo Chen}, \bibinfo{person}{Bing Bai}, {and} \bibinfo{person}{Fei Wang}.} \bibinfo{year}{2025}\natexlab{}.
\newblock \showarticletitle{Graph Evidential Learning for Anomaly Detection}. In \bibinfo{booktitle}{\emph{Proceedings of the 31st ACM SIGKDD Conference on Knowledge Discovery and Data Mining}}. \bibinfo{pages}{3122--3133}.
\newblock
\href{https://doi.org/10.1145/3711896.3736989}{doi:\nolinkurl{10.1145/3711896.3736989}}


\bibitem[Wesego(2025)]%
        {wesego2025ddgae}
\bibfield{author}{\bibinfo{person}{Daniel Wesego}.} \bibinfo{year}{2025}\natexlab{}.
\newblock \showarticletitle{Graph Representation Learning with Diffusion Generative Models}.
\newblock \bibinfo{journal}{\emph{arXiv preprint arXiv:2501.13133}} (\bibinfo{year}{2025}).
\newblock
\urldef\tempurl%
\url{https://arxiv.org/abs/2501.13133}
\showURL{%
\tempurl}


\bibitem[Wu et~al\mbox{.}(2021)]%
        {wu2021graphmixup}
\bibfield{author}{\bibinfo{person}{Lirong Wu}, \bibinfo{person}{Haitao Lin}, \bibinfo{person}{Zhangyang Gao}, \bibinfo{person}{Cheng Tan}, {and} \bibinfo{person}{Stan~Z. Li}.} \bibinfo{year}{2021}\natexlab{}.
\newblock \showarticletitle{{GraphMixup}: Improving Class-Imbalanced Node Classification on Graphs by Self-supervised Context Prediction}.
\newblock \bibinfo{journal}{\emph{arXiv preprint arXiv:2106.11133}} (\bibinfo{year}{2021}).
\newblock
\urldef\tempurl%
\url{https://arxiv.org/abs/2106.11133}
\showURL{%
\tempurl}


\bibitem[Xu et~al\mbox{.}(2020)]%
        {xu2020tgat}
\bibfield{author}{\bibinfo{person}{Da Xu}, \bibinfo{person}{Chuanwei Ruan}, \bibinfo{person}{Evren Korpeoglu}, \bibinfo{person}{Sushant Kumar}, {and} \bibinfo{person}{Kannan Achan}.} \bibinfo{year}{2020}\natexlab{}.
\newblock \showarticletitle{Inductive Representation Learning on Temporal Graphs}. In \bibinfo{booktitle}{\emph{International Conference on Learning Representations}}.
\newblock
\urldef\tempurl%
\url{https://openreview.net/forum?id=rJeW1yHYwH}
\showURL{%
\tempurl}


\bibitem[Yang et~al\mbox{.}(2023)]%
        {zhu2023ddm}
\bibfield{author}{\bibinfo{person}{Run Yang}, \bibinfo{person}{Yuling Yang}, \bibinfo{person}{Fan Zhou}, {and} \bibinfo{person}{Qiang Sun}.} \bibinfo{year}{2023}\natexlab{}.
\newblock \showarticletitle{Directional Diffusion Models for Graph Representation Learning}. In \bibinfo{booktitle}{\emph{Advances in Neural Information Processing Systems}}, Vol.~\bibinfo{volume}{36}.
\newblock
\urldef\tempurl%
\url{https://proceedings.neurips.cc/paper_files/paper/2023/hash/6751ee6546b31ceb7d4ee12276b9f4d9-Abstract-Conference.html}
\showURL{%
\tempurl}


\bibitem[Yu et~al\mbox{.}(2023)]%
        {yu2023dygformer}
\bibfield{author}{\bibinfo{person}{Le Yu}, \bibinfo{person}{Leilei Sun}, \bibinfo{person}{Bowen Du}, {and} \bibinfo{person}{Weifeng Lv}.} \bibinfo{year}{2023}\natexlab{}.
\newblock \showarticletitle{Towards Better Dynamic Graph Learning: New Architecture and Unified Library}. In \bibinfo{booktitle}{\emph{Advances in Neural Information Processing Systems}}, Vol.~\bibinfo{volume}{36}.
\newblock
\urldef\tempurl%
\url{https://proceedings.neurips.cc/paper_files/paper/2023/hash/d611019afba70d547bd595e8a4158f55-Abstract-Conference.html}
\showURL{%
\tempurl}


\bibitem[Yun et~al\mbox{.}(2022)]%
        {yun2022lte4g}
\bibfield{author}{\bibinfo{person}{Sukwon Yun}, \bibinfo{person}{Kibum Kim}, \bibinfo{person}{Kanghoon Yoon}, {and} \bibinfo{person}{Chanyoung Park}.} \bibinfo{year}{2022}\natexlab{}.
\newblock \showarticletitle{{LTE4G}: Long-Tail Experts for Graph Neural Networks}. In \bibinfo{booktitle}{\emph{Proceedings of the 31st ACM International Conference on Information and Knowledge Management}}. \bibinfo{pages}{2434--2443}.
\newblock
\href{https://doi.org/10.1145/3511808.3557381}{doi:\nolinkurl{10.1145/3511808.3557381}}


\bibitem[Zhao et~al\mbox{.}(2021)]%
        {zhao2021graphsmote}
\bibfield{author}{\bibinfo{person}{Tianxiang Zhao}, \bibinfo{person}{Xiang Zhang}, {and} \bibinfo{person}{Suhang Wang}.} \bibinfo{year}{2021}\natexlab{}.
\newblock \showarticletitle{{GraphSMOTE}: Imbalanced Node Classification on Graphs with Graph Neural Networks}. In \bibinfo{booktitle}{\emph{Proceedings of the 14th ACM International Conference on Web Search and Data Mining}}.
\newblock
\href{https://doi.org/10.1145/3437963.3441720}{doi:\nolinkurl{10.1145/3437963.3441720}}


\bibitem[Zheng et~al\mbox{.}(2025)]%
        {zheng2025dgmae}
\bibfield{author}{\bibinfo{person}{Ziyu Zheng}, \bibinfo{person}{Yaming Yang}, \bibinfo{person}{Ziyu Guan}, \bibinfo{person}{Wei Zhao}, {and} \bibinfo{person}{Weigang Lu}.} \bibinfo{year}{2025}\natexlab{}.
\newblock \showarticletitle{Discrepancy-Aware Graph Mask Auto-Encoder}. In \bibinfo{booktitle}{\emph{Proceedings of the 31st ACM SIGKDD Conference on Knowledge Discovery and Data Mining}}. \bibinfo{pages}{4038--4049}.
\newblock
\urldef\tempurl%
\url{https://arxiv.org/abs/2506.19343}
\showURL{%
\tempurl}


\bibitem[Zhou et~al\mbox{.}(2024)]%
        {zhou2024lgd}
\bibfield{author}{\bibinfo{person}{Cai Zhou}, \bibinfo{person}{Xiyuan Wang}, {and} \bibinfo{person}{Muhan Zhang}.} \bibinfo{year}{2024}\natexlab{}.
\newblock \showarticletitle{Unifying Generation and Prediction on Graphs with Latent Graph Diffusion}. In \bibinfo{booktitle}{\emph{Advances in Neural Information Processing Systems}}, Vol.~\bibinfo{volume}{37}.
\newblock
\urldef\tempurl%
\url{https://proceedings.neurips.cc/paper_files/paper/2024/hash/718d02a76d69686a36eccc8cde3e6a41-Abstract-Conference.html}
\showURL{%
\tempurl}


\bibitem[Zhou et~al\mbox{.}(2018)]%
        {zhou2018sparc}
\bibfield{author}{\bibinfo{person}{Dawei Zhou}, \bibinfo{person}{Jingrui He}, \bibinfo{person}{Hongxia Yang}, {and} \bibinfo{person}{Wei Fan}.} \bibinfo{year}{2018}\natexlab{}.
\newblock \showarticletitle{{SPARC}: Self-Paced Network Representation for Few-Shot Rare Category Characterization}. In \bibinfo{booktitle}{\emph{Proceedings of the 24th ACM SIGKDD International Conference on Knowledge Discovery and Data Mining}}.
\newblock
\href{https://doi.org/10.1145/3219819.3219952}{doi:\nolinkurl{10.1145/3219819.3219952}}


\bibitem[Zhou and Gong(2023)]%
        {zhou2023graphsr}
\bibfield{author}{\bibinfo{person}{Mengting Zhou} {and} \bibinfo{person}{Zhiguo Gong}.} \bibinfo{year}{2023}\natexlab{}.
\newblock \showarticletitle{{GraphSR}: A Data Augmentation Algorithm for Imbalanced Node Classification}. In \bibinfo{booktitle}{\emph{Proceedings of the AAAI Conference on Artificial Intelligence}}, Vol.~\bibinfo{volume}{37}. \bibinfo{pages}{4954--4962}.
\newblock
\href{https://doi.org/10.1609/aaai.v37i4.25622}{doi:\nolinkurl{10.1609/aaai.v37i4.25622}}


\bibitem[Zhu et~al\mbox{.}(2025)]%
        {zhu2025sdmg}
\bibfield{author}{\bibinfo{person}{Junyou Zhu}, \bibinfo{person}{Langzhou He}, \bibinfo{person}{Chao Gao}, \bibinfo{person}{Dongpeng Hou}, \bibinfo{person}{Zhen Su}, \bibinfo{person}{Philip~S. Yu}, \bibinfo{person}{Juergen Kurths}, {and} \bibinfo{person}{Frank Hellmann}.} \bibinfo{year}{2025}\natexlab{}.
\newblock \showarticletitle{{SDMG}: Smoothing Your Diffusion Models for Powerful Graph Representation Learning}. In \bibinfo{booktitle}{\emph{Proceedings of the 42nd International Conference on Machine Learning}} \emph{(\bibinfo{series}{Proceedings of Machine Learning Research}, Vol.~\bibinfo{volume}{267})}. \bibinfo{pages}{79815--79835}.
\newblock
\urldef\tempurl%
\url{https://proceedings.mlr.press/v267/zhu25g.html}
\showURL{%
\tempurl}


\end{thebibliography}

\end{document}